\documentclass[pdflatex,sn-mathphys-num]{sn-jnl}

\usepackage{graphicx}%
\usepackage{multirow}%
\usepackage{amsmath,amssymb,amsfonts}%
\usepackage{amsthm}%
\usepackage{mathrsfs}%
\usepackage[title]{appendix}%
\usepackage[table]{xcolor}%
\usepackage{textcomp}%
\usepackage{manyfoot}%
\usepackage{booktabs}%
\usepackage{algorithm}%
\usepackage{algorithmicx}%
\usepackage{algpseudocode}%
\usepackage{listings}%
\usepackage{adjustbox}
\usepackage{subcaption} 

\theoremstyle{thmstyleone}%
\newtheorem{theorem}{Theorem}
\newtheorem{lemma}[theorem]{Lemma}%

\theoremstyle{thmstyletwo}%

\theoremstyle{thmstylethree}%

\begin{document}

\title[Article Title]{CFNN: Continued Fraction Neural Network}


\author*[1,2]{\fnm{Chao} \sur{Wang}}\email{cwang@shu.edu.cn}

\author[1,2]{\fnm{Xuancheng} \sur{Zhou}}\email{xuanchengz@shu.edu.cn}

\author[3]{\fnm{Ruilin} \sur{Hou}}\email{hr0962592@gmail.com}

\author[4]{\fnm{Xiaoyu} \sur{Cheng}}\email{2036974785@shu.edu.cn}

\author[5]{\fnm{Ruiyi} \sur{Ding}}\email{dingruiyi@shu.edu.cn}

\nocite{*}

\affil*[1]{\orgdiv{School of Future Technology}, \orgname{Shanghai University}, \orgaddress{\city{Shanghai}, \postcode{200444}, \country{China}}}

\affil[2]{\orgdiv{Institute of Artificial Intelligence}, \orgname{Shanghai University}, \orgaddress{\city{Shanghai}, \postcode{200444}, \country{China}}}

\affil[3]{\orgdiv{School of Computer Science and Technology}, \orgname{Shanghai University}, \orgaddress{\city{Shanghai}, \postcode{200444}, \country{China}}}

\affil[4]{\orgdiv{College of Sciences}, \orgname{Shanghai University}, \orgaddress{\city{Shanghai}, \postcode{200444}, \country{China}}}

\affil[5]{\orgdiv{SILC Business School}, \orgname{Shanghai University}, \orgaddress{\city{Shanghai}, \postcode{200444}, \country{China}}}


\abstract{The accurate characterization of non-linear functional manifolds with local singularities and sharp transitions is a fundamental challenge in scientific computing. While Multi-Layer Perceptrons (MLPs) dominate function approximation, their inherent polynomial inductive bias leads to a pronounced ``spectral bias'', causing failure in resolving high-curvature features without prohibitive parameter scaling. Here, we introduce Continued Fraction Neural Networks (CFNNs), a novel architectural family that transcends these limitations by integrating the rigorous representation efficiency of continued fractions with gradient-based optimization. By parameterizing fractional terms as projection-based scalar polynomial transformations, CFNNs provide a ``rational inductive bias'' that aligns with Padé approximation theory, enabling the CFNN family to capture complex asymptotic behaviors and discontinuities with extreme parameter frugality. We provide formal approximation bounds demonstrating exponential convergence rates with respect to model depth and polynomial degree, along with rigorous stability guarantees for the proposed rational parameterizations. To overcome the recursive instability of deep fractional structures, we develop three specialized implementations: CFNN-Boost, which utilizes stage-wise additive modeling; CFNN-MoE, which partitions the input manifold using anisotropic radial basis function gated experts for adaptive feature space weighting; and CFNN-Hybrid, which combines parallel rational units with skip connections and provably bounded gradients. Benchmark results across diverse mathematical manifolds and real-world physical modeling tasks demonstrate that CFNNs consistently outperform MLPs in precision while requiring one to two orders of magnitude fewer parameters. Notably, CFNNs exhibit up to a 47-fold improvement in noise robustness and produce feature attributions consistent with underlying physical laws. By bridging the gap between black-box flexibility and white-box transparency, the CFNN family establishes a reliable ``grey-box'' paradigm for high-precision modeling in AI-driven scientific research.}

\keywords{Continued Fraction, Rational Inductive Bias, Grey-box Modeling}



\maketitle

\section{Introduction}\label{sec: Introduction}

Function approximation stands as a cornerstone of modern scientific discovery, spanning disciplines from computational physics to complex dynamical system modeling \cite{SpectralBiasIJCAI2021}. At its core, the challenge lies in identifying mathematical representations capable of accurately characterizing high-dimensional, nonlinear mappings \cite{UniversalApproximate}. While Multi-Layer Perceptrons (MLPs) have achieved remarkable success in modeling unstructured data, their application to precision scientific computing reveals fundamental structural limitations \cite{MLP!}. Most notably, standard neural architectures exhibit a pronounced ``spectral bias'', effectively capturing low-frequency global trends but struggling to resolve local near-singularities, sharp transitions, or intrinsic rational structures \cite{SpectralBiasICML2019,SpectralBiasNIPS2024}. Conventional strategies to mitigate these deficiencies frequently incur substantial computational costs, leading to parameter redundancy, numerical instability, and a lack of physical interpretability critical for scientific validation \cite{SpectralBiasAAAI2022,SpectralBiasNature2021}.

An alternative paradigm for function representation, with deep roots in classical numerical analysis, is the theory of Continued Fractions \cite{CFBook1980william}. Distinguished by their nested fractional structure and profound connection to Padé approximation theory, continued fractions offer unparalleled efficiency in modeling both local singularities and global asymptotic behaviors with a minimal set of parameters \cite{PadeApproximateJAT1997,RationalNNNIPS2020}. However, the seamless integration of continued fractions into modern gradient-based optimization frameworks has remained an elusive goal \cite{GRUGradientProblemIFAC2020}. Their deeply recursive nonlinear nature presents formidable challenges, including severe gradient instability and significant parallelization bottlenecks on modern hardware, which have historically restricted their adoption in deep learning \cite{NumCFInterpolationUMJ2024}.

To bridge this gap, we introduce the Continued Fraction Neural Network (CFNN) family, a novel architectural framework designed to synthesize the robust representation capabilities of continued fractions with the flexibility of contemporary deep learning. We propose a systematic parameterization that models fractional terms as projection-based scalar polynomial transformations, thereby ensuring analytical tractability and structural transparency. Crucially, we establish formal approximation bounds demonstrating that CFNN approximation errors decay exponentially with respect to the product of continued fraction depth and polynomial degree, in stark contrast to the polynomial decay rates of standard MLPs. Furthermore, we provide rigorous stability analysis characterizing the gradient boundedness and Lipschitz continuity properties of our regularized rational parameterizations. To address the optimization challenges inherent in deep recursion, we develop three specialized variants tailored to distinct topological challenges: CFNN-Boost employs stage-wise additive modeling to directly capture functional discontinuities and abrupt transitions through residual-based refinement; CFNN-MoE achieves adaptive isolation of local singularities via anisotropic RBF-gated expert partitioning, enabling flexible approximation of heterogeneous manifolds with varying degrees of regularity through learnable feature space weighting; and CFNN-Hybrid integrates parallel rational units with regularized denominators to efficiently fit continuous manifolds exhibiting extreme curvature while maintaining absolute numerical stability and provably bounded gradients across the real domain.

Through a multi-dimensional evaluation encompassing special mathematical functions and diverse real-world modeling tasks, we demonstrate that the CFNN family consistently achieves superior approximation accuracy while requiring one to two orders of magnitude fewer parameters than calibrated MLPs. Furthermore, CFNNs exhibit a remarkable 47-fold improvement in noise suppression and extract feature hierarchies highly consistent with underlying physical laws. By balancing extreme expressivity with ``grey-box'' transparency, the CFNN family provides a powerful and interpretable modeling tool for the next generation of AI-driven scientific research.

\section{Results}\label{sec: Results}

\subsection{Optimization landscapes and structural scalability challenges}\label{subsec: Scalability}

\begin{figure}[h]
  \centering
    \includegraphics[width=\linewidth]{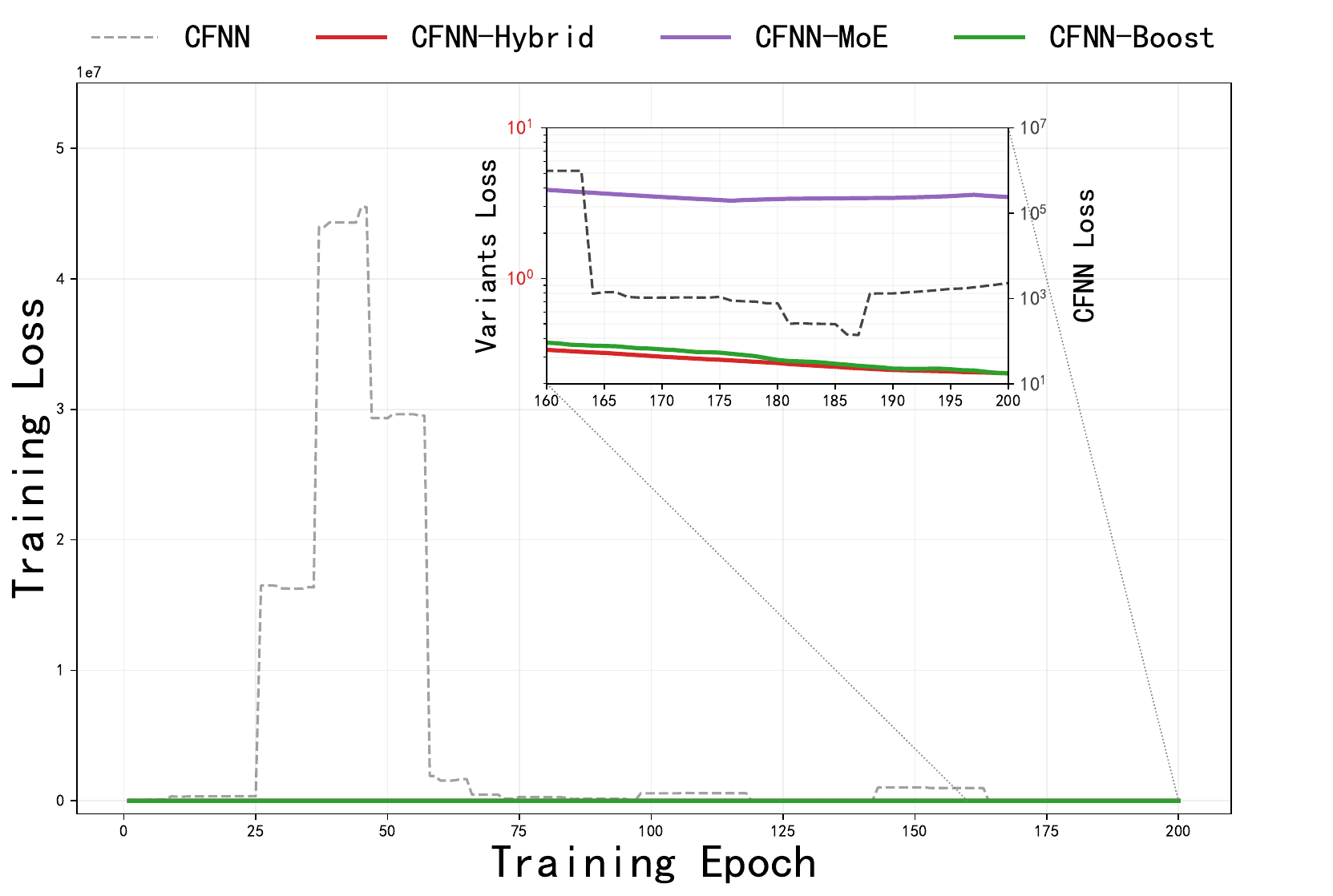} 
  \caption{Training-loss trajectories for the foundational CFNN and its three stabilized variants, measured by mean squared error. The dashed CFNN curve exhibits severe optimization instability, with abrupt spikes and extended oscillatory plateaus that reach the $10^{7}$ scale. In contrast, CFNN-Hybrid and CFNN-Boost descend smoothly to the lowest final losses, while CFNN-MoE remains stable but converges to a higher error floor. The inset magnifies the late training stage (epochs 160--200), making the separation among the three stable variants explicit.}
  \label{fig:train_loss_stability}
\end{figure}

The representational potency of CFNNs originates from their nested recursive structures, which mirror the exceptional function approximation capabilities of classical continued fractions. However, this recursive depth introduces significant optimization hurdles, specifically manifesting as severe gradient instability during training \cite{GRUGradientProblemIFAC2020,NumCFInterpolationUMJ2024}. This instability exacerbates as the network depth increases; as illustrated in Fig. \ref{fig:train_loss_stability}, the foundational CFNN architecture exhibits abrupt loss spikes and prolonged oscillatory plateaus, with training error rising to the $10^{7}$ scale before partially collapsing. By contrast, CFNN-Hybrid and CFNN-Boost decrease smoothly throughout training, while CFNN-MoE remains stable but converges to a visibly higher loss level in the zoomed late-epoch view. Such stochasticity in the optimization landscape fundamentally constrains the model's ability to achieve high-precision functional fitting \cite{GradientDisappearanceAITCS2025}.

\begin{figure}[h]
  \centering
  \includegraphics[width=\linewidth]{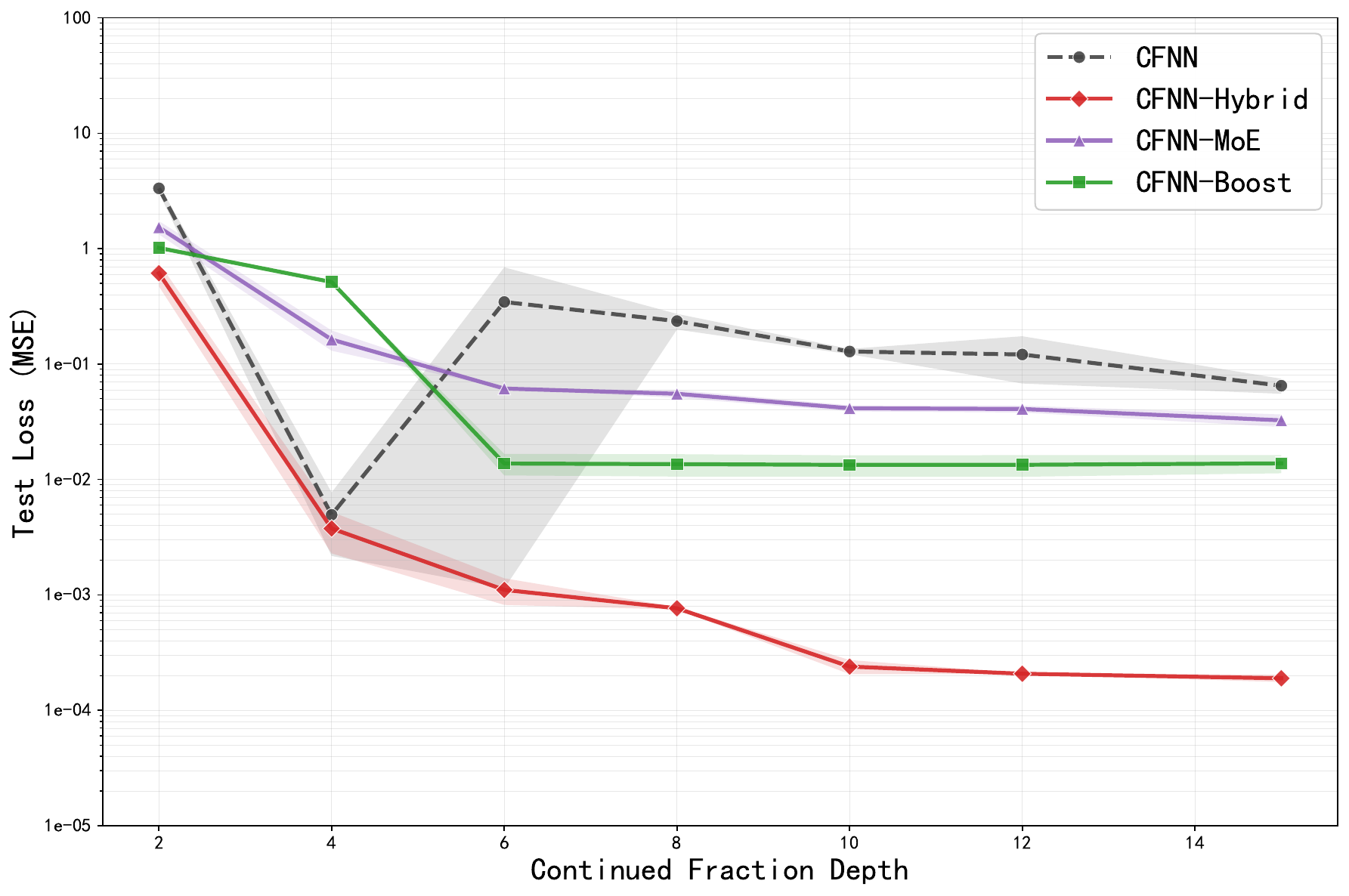}
  \caption{Test-loss scaling with continued-fraction depth for the foundational CFNN and its three stabilized variants, measured by mean squared error on a logarithmic scale. The foundational CFNN exhibits strongly non-monotonic behaviour: performance improves sharply from depth 2 to 4, deteriorates at depth 6, and only partially recovers at larger depths. In contrast, CFNN-Hybrid shows the most favorable scaling trend, with test loss decreasing consistently to the $10^{-4}$ regime as depth increases. CFNN-MoE and CFNN-Boost also improve more smoothly than the foundational CFNN, but converge to higher loss plateaus in the $10^{-2}$ to $10^{-1}$ range.}
  \label{fig:parameter_scaling}
\end{figure}

Theoretical frameworks suggest that increasing the depth of a continued fraction should monotonically enhance its representational capacity \cite{CFBook2018Wall,ConvergenceTheoremofCF1940TAMS}. Paradoxically, our experiments challenge this conventional wisdom in the context of neural optimization. By evaluating CFNNs of varying depths while maintaining a constant polynomial order, we observed a critical performance regression. Notably, the foundational CFNN reaches its best performance around depth 4, but then deteriorates sharply at depth 6 and only partially recovers thereafter, as shown in Fig. \ref{fig:parameter_scaling}. These phenomena suggest a structural limitation inherent in the basic CFNN formulation: its expressive power is highly sensitive to the specific recursive topology. The accumulation of gradient stochasticity across deeper layers creates a convoluted optimization manifold, leading to a profound divergence between theoretical representational limits and actual empirical performance.

\begin{figure}[h]
  \centering
  \includegraphics[width=\linewidth]{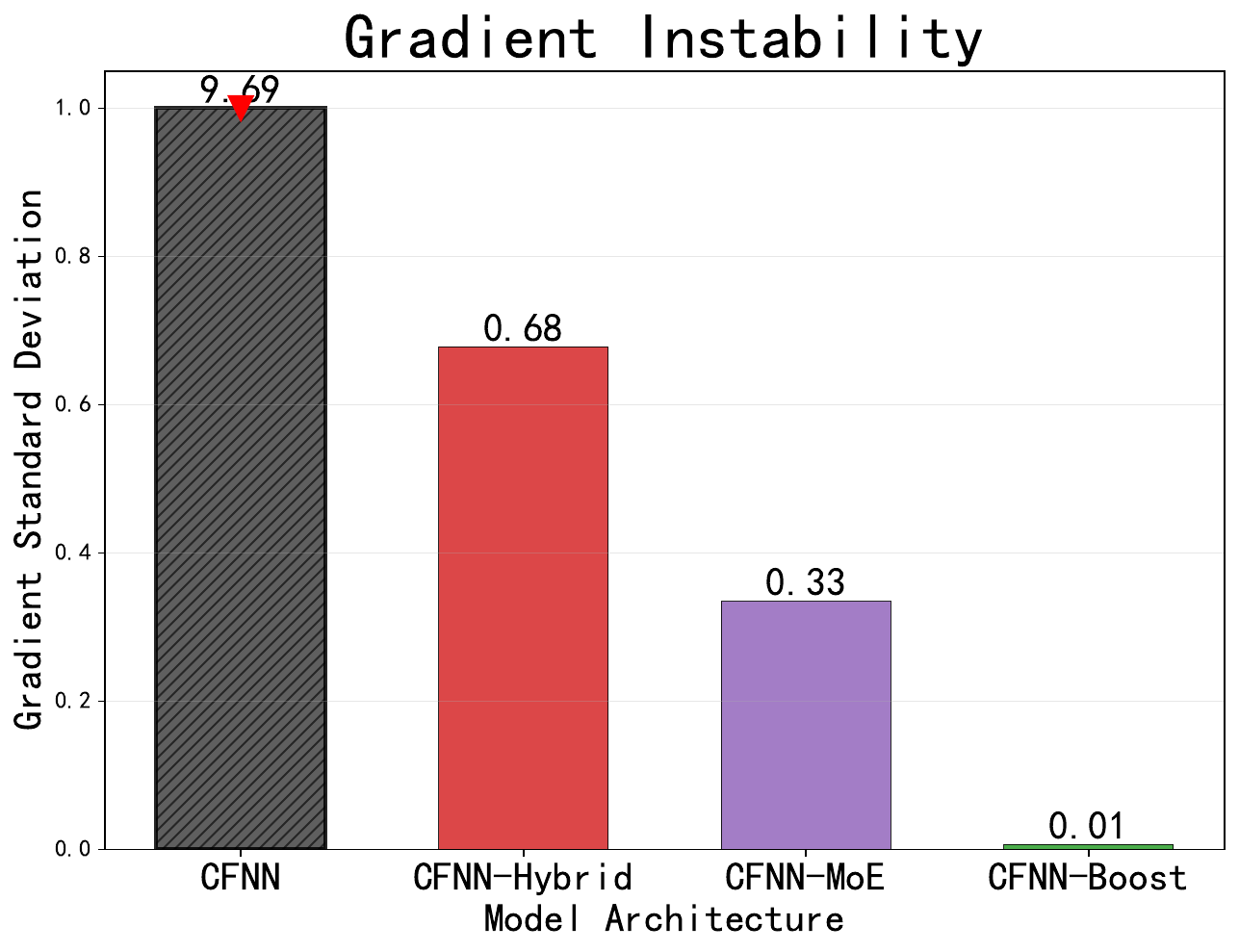}
  \caption{Comparison of gradient standard deviation across the foundational CFNN and its three stabilized variants. Lower values indicate greater optimization stability. The foundational CFNN exhibits an extreme gradient fluctuation of 9.69, whereas CFNN-Hybrid, CFNN-MoE, and CFNN-Boost reduce this statistic to 0.68, 0.33, and 0.01, respectively. The CFNN bar is truncated at 1 for visual clarity, with its true value annotated above the bar. These results provide a quantitative explanation for the severe loss oscillations in Fig. \ref{fig:train_loss_stability}.}
  \label{fig:train_gradient_std}
\end{figure}

To address these bottlenecks, we constructed and benchmarked three architectural variants: CFNN-Boost, CFNN-MoE, and CFNN-Hybrid, designed to preserve the rational inductive bias while enhancing learning stability. These variants demonstrate a marked improvement in gradient regularity compared to the foundational CFNN. As shown in Fig. \ref{fig:train_gradient_std}, the foundational CFNN exhibits by far the largest gradient fluctuation, with a standard deviation of 9.69, whereas CFNN-Hybrid, CFNN-MoE, and CFNN-Boost reduce this value to 0.68, 0.33, and 0.01, respectively. This ordering is consistent with the smoother training trajectories observed in Fig. \ref{fig:train_loss_stability}, although the lowest gradient variance does not necessarily imply the lowest final fitting error. Crucially, these variants also exhibit robust parameter scalability. As illustrated in Fig. \ref{fig:parameter_scaling}, CFNN-Hybrid shows the strongest and most persistent improvement with depth, ultimately reaching the lowest test loss across the full range. CFNN-MoE and CFNN-Boost also improve far more smoothly than the foundational CFNN, but both flatten into higher error plateaus at larger depths. It is evident that while the performance of the foundational CFNN exhibits non-negligible variability, all three proposed variants consistently maintain stable performance. These results establish a robust foundation for the deployment of the CFNN variants family in complex, large-scale modeling tasks.

\subsection{Learning efficacy across diverse functional manifolds}\label{subsec: LearningEfficacy}

The foundational learning characteristics of the CFNN architectural family were initially evaluated through synthetic datasets comprising rational and sharp-transition functions. Utilizing RMSE and the Lead Metric defined in Equation \ref{eq:lead_metric} to benchmark performance against standard MLPs, we observe distinct advantages in the representational efficacy of CFNN variants. In comparative experiments conducted at a calibrated scale of approximately 200 learnable parameters, all three specialized CFNN variants outperform the MLP baseline by substantial margins across the four test functions, as summarized in Table \ref{tab:basic_function_fit_exp}. Among these variants, CFNN-MoE and CFNN-Hybrid deliver the strongest overall performance, while CFNN-Boost achieves the lowest error on the Runge function.

We also benchmark these results against Kolmogorov-Arnold Networks (KAN) \cite{KAN!}, another prominent representative of the MLP-alternative paradigm. While KAN achieves the highest absolute precision in these synthetic fitting tasks, it is important to view CFNN and KAN as two distinct yet complementary methodologies targeting the same fundamental goal of transcending MLP limitations. KAN’s exceptional accuracy in this regime is largely due to its spline-based inductive bias, which is highly optimized for numerical precision in smooth functional manifolds. In contrast, the CFNN family leverages a rational inductive bias derived from continued fractions. As detailed in subsequent sections, this performance divergence in idealized fitting is a natural consequence of their differing mathematical foundations, with CFNN offering unique strengths in noise robustness and physical consistency where pure fitting precision may be secondary to structural resilience.

\begin{figure}[h]
  \centering
  \begin{subfigure}[h]{0.9\textwidth}
    \centering
    \includegraphics[width=\linewidth]{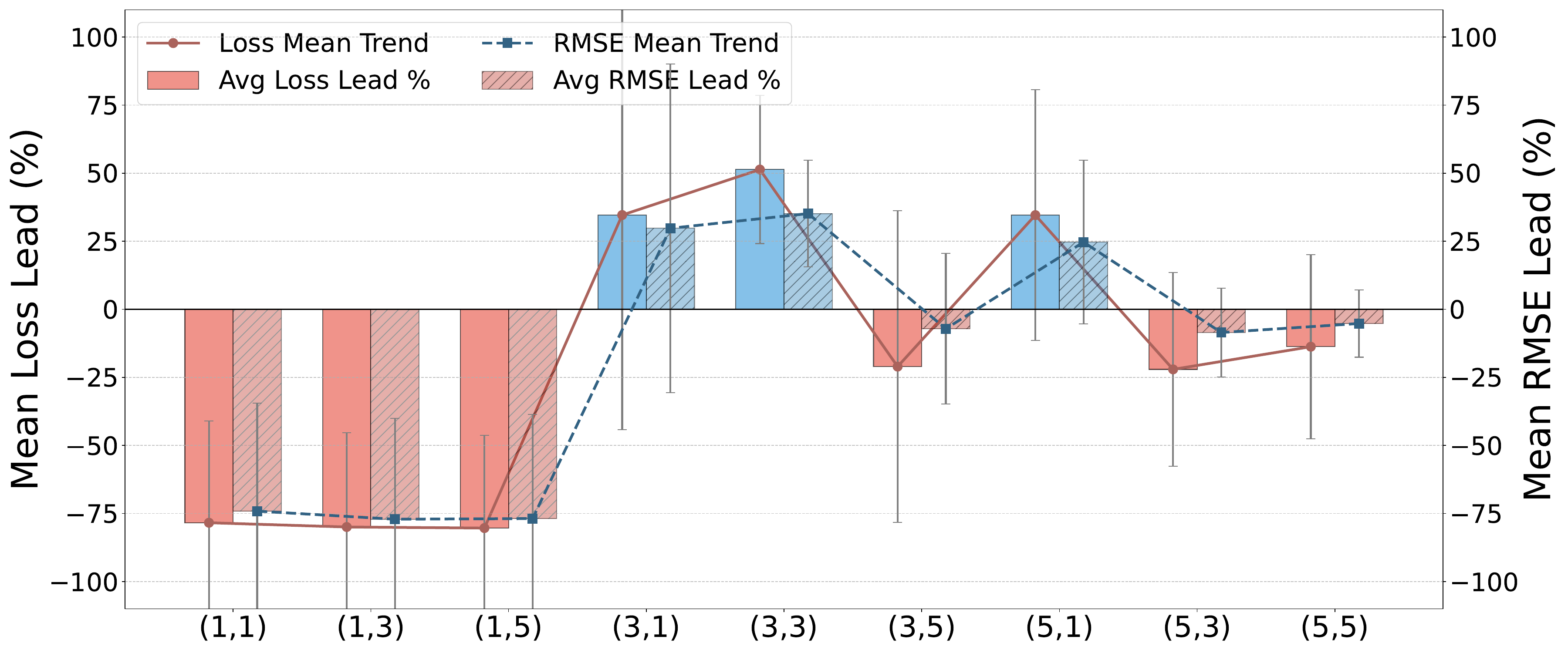}
    \caption{Lead Metric of CFNN}
    \label{figs:lead_cfnn}
  \end{subfigure}
  \hfill
  \begin{subfigure}[h]{0.9\textwidth}    
    \centering
    \includegraphics[width=\linewidth]{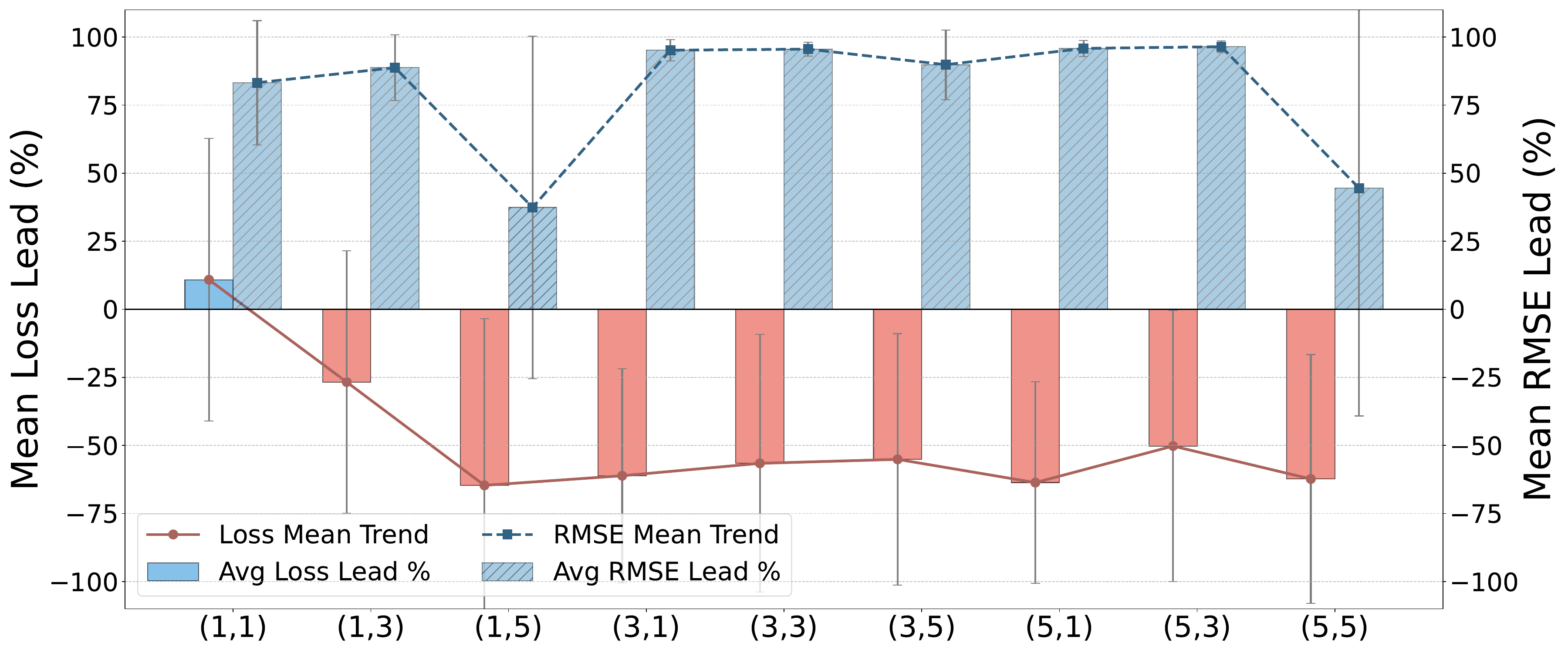}
    \caption{Lead Metric of CFNN-Boost}
    \label{fig:lead_boost}
  \end{subfigure}
  \begin{subfigure}[h]{0.9\textwidth}
    \centering
    \includegraphics[width=\linewidth]{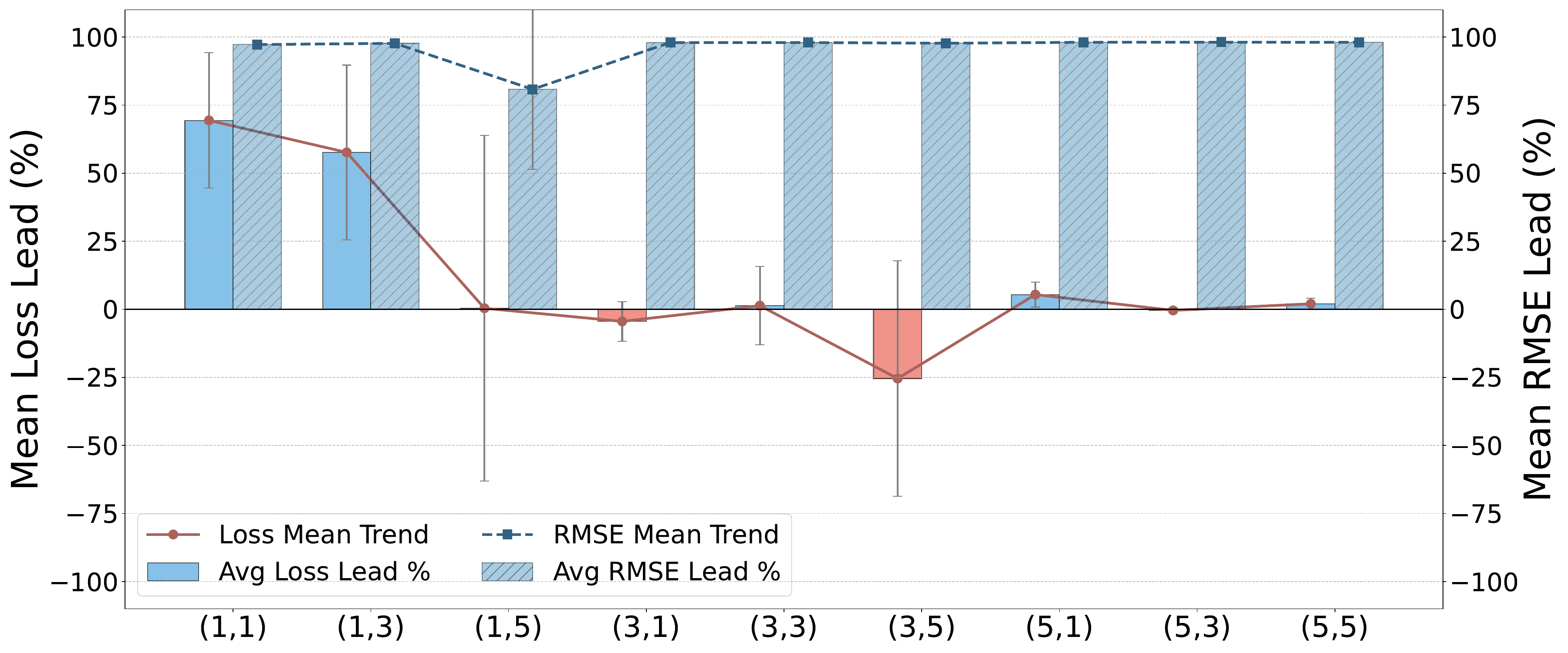}
    \caption{Lead Metric of CFNN-MoE}
    \label{fig:lead_moe}
  \end{subfigure}
  \hfill
  \begin{subfigure}[h]{0.9\textwidth}
    \centering
    \includegraphics[width=\linewidth]{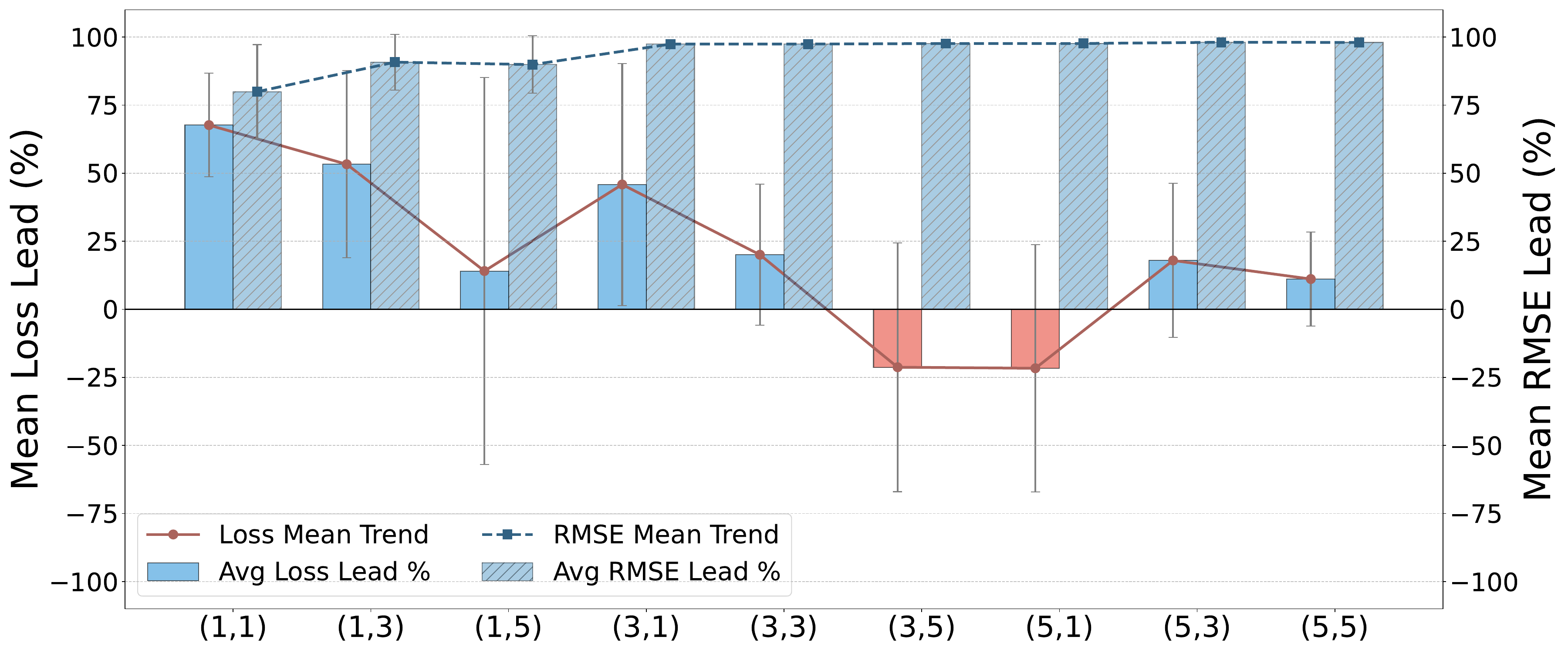}
    \caption{Lead Metric of CFNN-Hybrid}
    \label{fig:lead_hybrid}
  \end{subfigure}
  \caption{Lead-metric comparison between CFNN-family architectures and matched MLP baselines across structural configurations. Positive values indicate improvement over the corresponding MLP. The foundational CFNN shows strongly configuration-dependent behaviour, whereas CFNN-Boost, CFNN-MoE, and CFNN-Hybrid maintain broadly positive RMSE lead across most settings. Error bars denote variability across benchmark functions.}
  \label{figs:lead_cfnn_series}
\end{figure}

The superiority of CFNN-MoE is primarily attributed to its error-driven expert training strategy, which enables the model to sensitively capture local functional mutations while maintaining global stability. Conversely, CFNN-Hybrid achieves stable and high-precision fitting by integrating linear skip connections that balance global linear trends with nonlinear rational residuals. Furthermore, systematic ablation studies across varying model scales reveal that the foundational CFNN improves over the MLP baseline only in a limited subset of architectural configurations, whereas the three specialized variants maintain much stronger and more consistent positive RMSE lead across almost the entire grid, as shown in Fig. \ref{figs:lead_cfnn_series}. Notably, the CFNN-Boost variant realizes a substantial lead in test-set RMSE despite often exhibiting worse training losses than MLPs. This discrepancy highlights the implicit regularization capability inherent in its stage-wise additive modeling, which bolsters generalization against overfitting at the cost of higher prediction variance. In contrast, CFNN-MoE and CFNN-Hybrid sustain near-uniform RMSE gains across nearly all configurations, while their training-loss lead remains predominantly positive or close to neutral except for a few isolated settings. Taken together, these results indicate that the specialized CFNN variants convert the rational inductive bias into a much more stable fitting paradigm than the foundational architecture.

\subsubsection{Frequency-Agnostic Approximation and Spectral Bias Mitigation}

To rigorously evaluate the mitigation of spectral bias, we benchmarked the CFNN architectural family against standard MLPs and strong baselines explicitly optimized for high-frequency resolution, including Sinusoidal Representation Networks (SIREN), Random Fourier Features (RFF-MLP), and Chebyshev-KAN. We evaluated these models on a highly non-linear target manifold, $y = (x_1 x_2) / x_3$, utilizing Relative Power Spectral Density (Relative PSD) to normalize the residual energy against the target signal's inherent frequency decay.

\begin{figure}
  \centering
  \includegraphics[width=\linewidth]{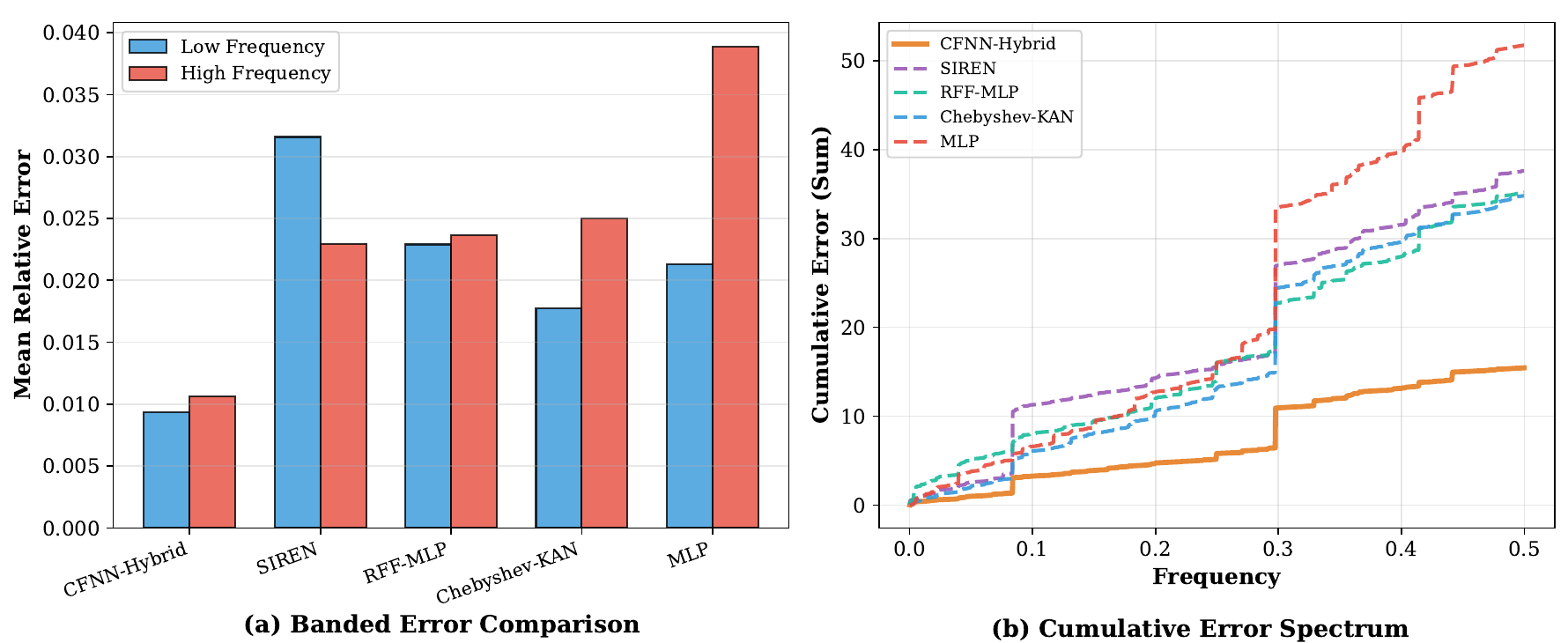}
  \caption{Frequency-domain comparison of approximation residuals for CFNN-Hybrid and competing baselines. (a) Mean relative error in low-frequency ($f < 0.25 f_{\max}$) and high-frequency ($f \ge 0.25 f_{\max}$) bands. MLP exhibits pronounced spectral bias, with high-frequency error ($0.0389$) substantially exceeding low-frequency error ($0.0213$), whereas SIREN displays the opposite imbalance, achieving lower high-frequency error ($0.0229$) at the cost of the worst low-frequency error ($0.0316$). CFNN-Hybrid attains the lowest error in both bands (0.0094 and 0.0106), indicating the most balanced frequency response. (b) Cumulative error spectrum across normalized frequency. CFNN-Hybrid maintains the lowest cumulative error throughout the full frequency range, while MLP accumulates the largest error and the other high-frequency-aware baselines remain intermediate.}
  \label{fig:freq_bias}
\end{figure}

Quantitative frequency-domain analysis reveals a fundamental trade-off in existing architectures, as shown in the banded error panel of Fig.~\ref{fig:freq_bias}a. The standard MLP exhibits severe spectral bias, suffering a high-frequency mean relative error of $0.0389$, which is $1.83\times$ higher than its low-frequency error ($0.0213$). Conversely, high-frequency-specialized models such as SIREN suppress high-frequency error more effectively ($0.0229$) but incur the worst low-frequency error among all methods ($0.0316$). RFF-MLP and Chebyshev-KAN partially balance these regimes, yet both remain substantially less uniform than CFNN-Hybrid. Strikingly, CFNN-Hybrid achieves the lowest absolute errors in both bands (low-frequency: $0.0094$, high-frequency: $0.0106$), indicating the most balanced behavior across the spectrum. The cumulative error spectrum in Fig. \ref{fig:freq_bias}b further corroborates this result: the CFNN-Hybrid curve remains the lowest over the entire frequency range and increases more gradually than all competing baselines, confirming superior global frequency-agnostic stability.

\begin{figure}
  \centering
  \includegraphics[width=\linewidth]{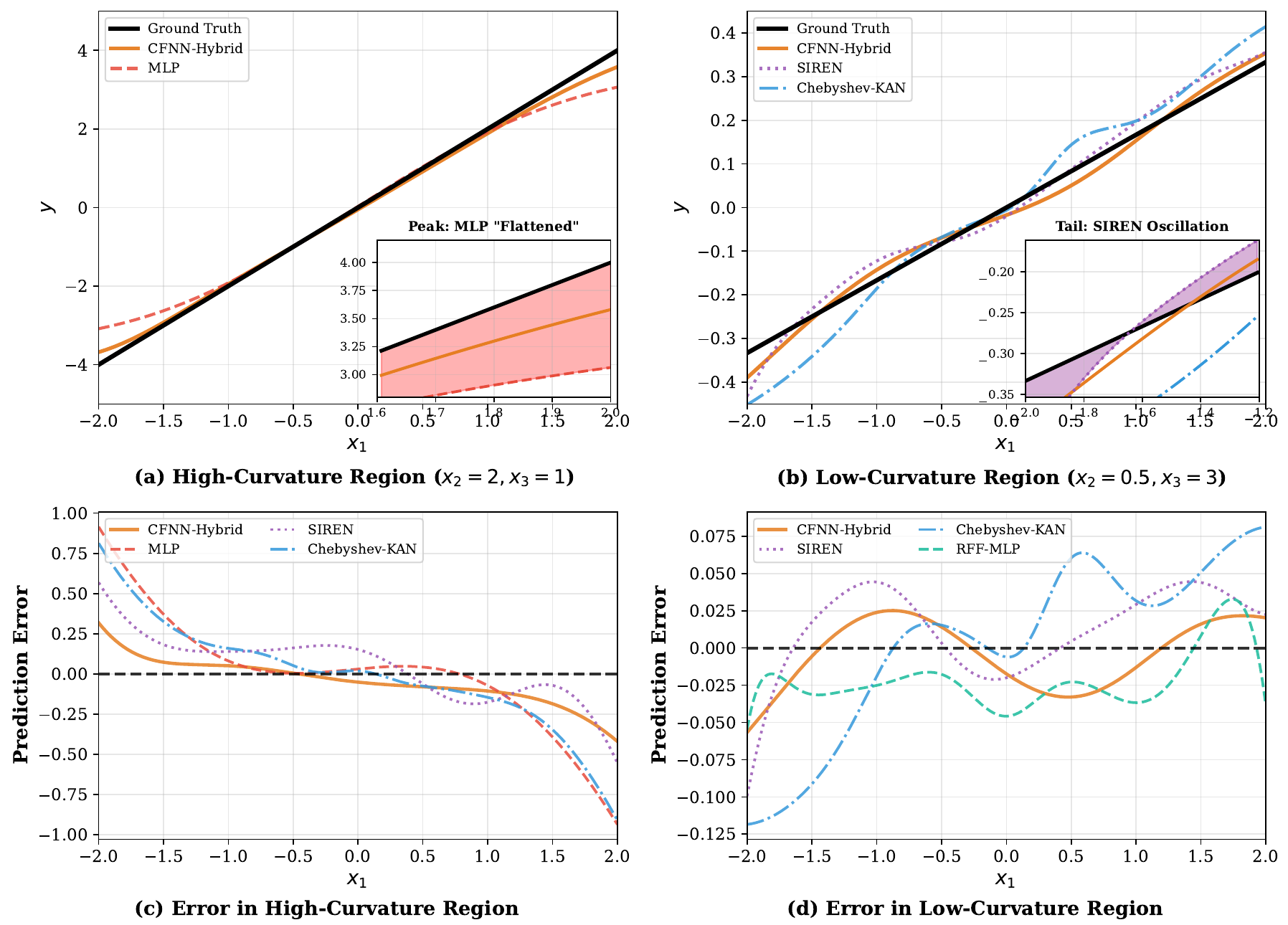}
  \caption{Spatial manifestation of spectral bias in function approximation. Top row: Cross-sectional fitting on high-curvature (left, $x_2=2, x_3=1$) and low-curvature (right, $x_2=0.5, x_3=3$) slices. Insets highlight critical failure modes: MLP exhibits peak flattening in high-curvature regions, while SIREN and Chebyshev-KAN introduce spurious oscillations in smooth regions. Bottom row: Residual error distributions. CFNN-Hybrid maintains near-zero error across both regimes, demonstrating adaptive resolution without trade-offs.}
  \label{fig:freq_bias_spatial}
\end{figure}

These spectral properties directly dictate the models' spatial fitting capabilities, as illustrated in Fig.~\ref{fig:freq_bias_spatial}. In the spatial domain, we analyzed cross-sectional topological slices of the manifold. In high-curvature regions (e.g., $x_2=2, x_3=1$), the MLP's inability to resolve high frequencies results in a catastrophic ``flattening'' of the sharp peak. In contrast, in low-curvature smooth regions (e.g., $x_2=0.5, x_3=3$), models with global high-frequency bases, such as SIREN and Chebyshev-KAN, introduce spurious structural oscillations (Gibbs-like phenomena) into the functional tail. By leveraging its rational inductive bias, CFNN-Hybrid is the only architecture capable of achieving adaptive resolution—perfectly matching local sharp transitions while preserving absolute smoothness in flat continuous domains.

\subsection{Parameter frugality and intrinsic regularization against high-dimensional noise}

\begin{figure}
  \centering
  \includegraphics[width=\linewidth]{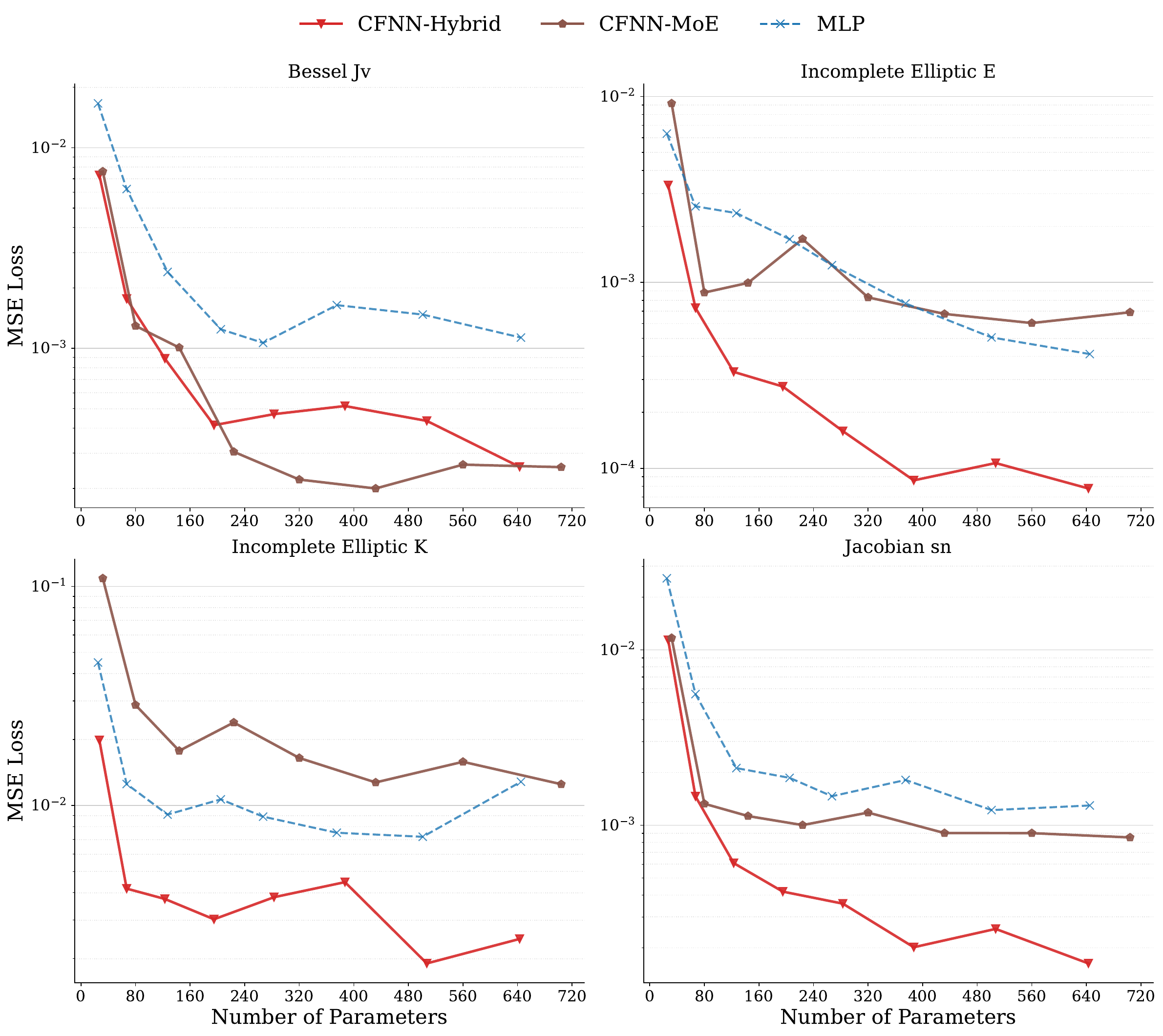}
  \caption{Pareto-style comparison of parameter efficiency on four challenging analytical function families. Each panel reports test MSE as a function of parameter count, so curves closer to the lower-left indicate better accuracy-efficiency trade-offs. Across all four tasks, CFNN-Hybrid forms the strongest frontier, maintaining lower error than the MLP baseline over nearly the full parameter range and continuing to improve where the MLP begins to saturate. CFNN-MoE also remains competitive and typically outperforms MLP at moderate-to-large scales, but generally plateaus above CFNN-Hybrid.}
  \label{fig:5func_parero}
\end{figure}

\begin{figure}
  \centering
  \includegraphics[width=\linewidth]{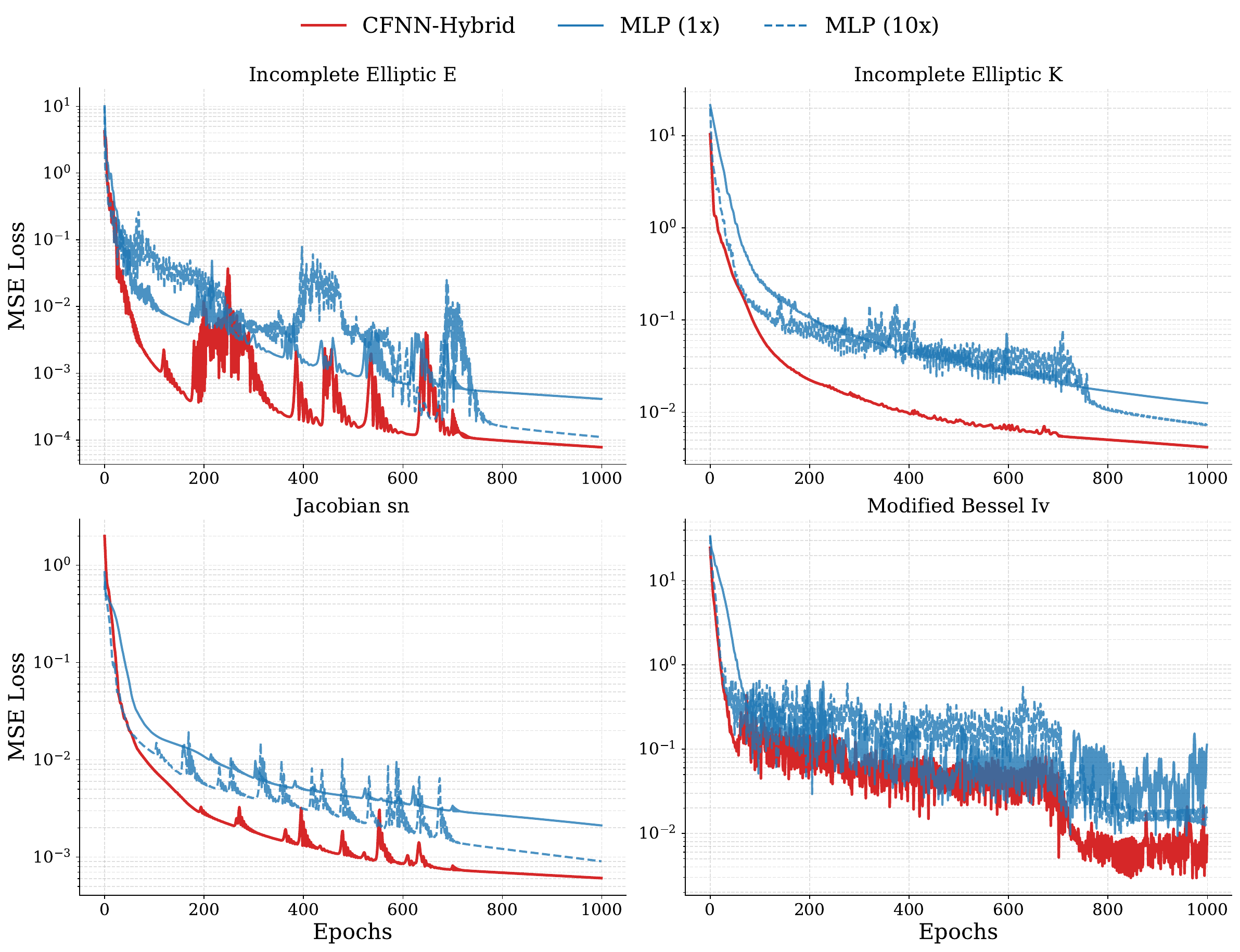}
  \caption{Training-loss dynamics of CFNN-Hybrid compared with matched-capacity MLP and a tenfold-wider MLP baseline across four analytical benchmark functions. Lower curves indicate faster convergence and better optimization performance. CFNN-Hybrid reaches low-error regimes earlier than both MLP baselines on all four tasks and maintains the lowest loss for most of training, despite the presence of occasional transient spikes on the more difficult manifolds. The enlarged MLP (10x) improves over the matched-scale MLP but typically remains above CFNN-Hybrid, indicating that the observed gain is not explained by parameter count alone.}
  \label{fig:training_dynamic}
\end{figure}

To rigorously evaluate the parameter efficiency of the CFNN architecture, we conducted a systematic Pareto frontier analysis across four complex functional manifolds: Jacobian elliptic functions, incomplete elliptic integrals of the first and second kind, and modified Bessel functions. By mapping the trade-off between model complexity and approximation precision, we observe that the representative CFNN variants shown in Fig. \ref{fig:5func_parero}, especially CFNN-Hybrid, establish a superior efficiency frontier compared to standard MLPs across nearly all parameter scales. Notably, even as the MLP performance begins to saturate, the CFNN-Hybrid variant maintains a steeper downward trend in loss on all four tasks, while CFNN-MoE also remains competitive over much of the parameter range. This behavior underscores the synergistic capacity improvement realized by synthesizing linear skip connections with the rational inductive bias of continued fractions. Further investigation into learning kinetics reveals that CFNN-Hybrid reaches low-error regimes substantially earlier than both matched-scale MLPs and enlarged MLP baselines. As shown in Fig. \ref{fig:training_dynamic}, it maintains lower training loss than both MLP (1x) and MLP (10x) across most of the optimization trajectory, establishing a substantial performance gap within the early and intermediate training epochs.

\begin{figure}
  \centering
  \includegraphics[width=\linewidth]{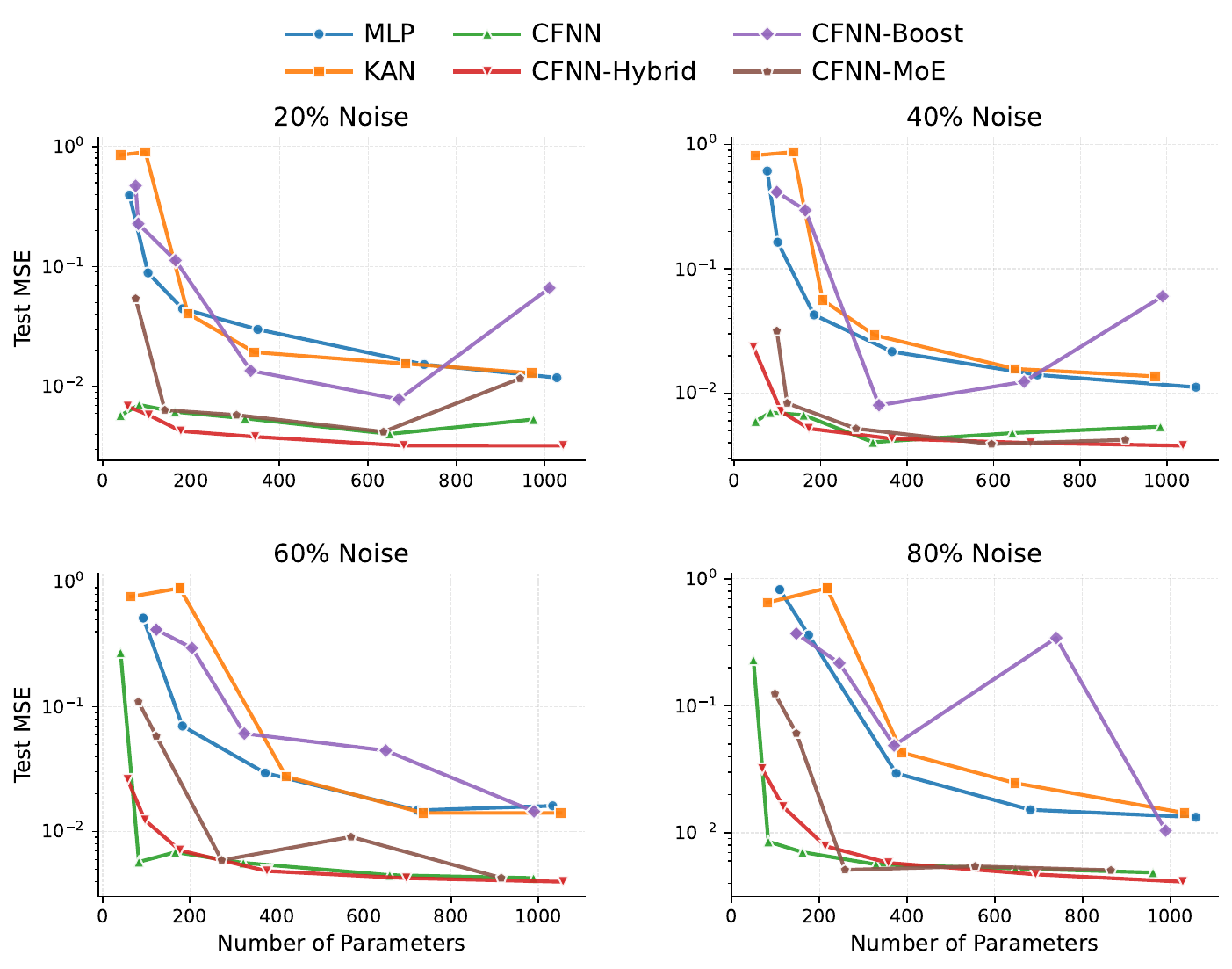}
  \caption{Pareto-style comparison of parameter efficiency under increasing noise-feature interference. Each panel reports test MSE as a function of parameter count for a fixed noise ratio, with lower curves indicating better robustness-efficiency trade-offs. Across the 20\%--80\% noise regimes shown here, CFNN-Hybrid remains on or near the strongest frontier and is closely followed by the foundational CFNN, whereas MLP and KAN generally require substantially more parameters to approach comparable error levels. CFNN-MoE remains competitive at moderate scales, while CFNN-Boost exhibits more configuration-dependent behaviour.}
  \label{fig:noise_pareto}
\end{figure}

This ``parameter-frugal'' nature provides more than just computational economy; it confers structural resilience against stochastic data corruption. Our systematic analysis spanning noise ratios from 20\% to 90\% reveals that the CFNN variants, specifically CFNN-Hybrid, consistently dominate the performance-complexity trade-off space. Within the 20\%--80\% noise regimes visualized in Fig. \ref{fig:noise_pareto}, CFNN-Hybrid remains on or near the strongest frontier across almost the full parameter range, while the foundational CFNN also stays highly competitive. This stands in stark contrast to KAN, which exhibits a much steeper efficiency penalty as noise increases, likely due to the vulnerability of its spline-based basis functions to local data irregularities. The full 20\%--90\% threshold analysis in Table \ref{tab:noise_threshold_params} further confirms that CFNN-Hybrid requires substantially fewer parameters than both MLP and KAN to achieve the same high-precision target.

Table \ref{tab:noise_threshold_params} further quantifies the architectural superiority of the CFNN family by examining the parameter overhead required to reach high-precision fitting thresholds. To achieve a convergence threshold of $MSE < 0.01$, MLPs require between 2 and 24 times more parameters than CFNN-Hybrid across the evaluated noise regimes, while KAN requires 8 to 24 times more parameters whenever it reaches the threshold at all. This efficiency gap highlights that the recursive fractional structure enables the network to recover the global functional hierarchy using a minimal set of learnable coefficients, whereas standard architectures must rely on massive over-parameterization to filter out stochastic interference. These results suggest that the rational structure of CFNN acts as an intrinsic regularization mechanism, suppressing deceptive features while amplifying physically consistent signals. Consequently, the CFNN family establishes a robust ``grey-box'' paradigm that reconciles high-precision approximation with rigorous structural resilience in adversarial data environments.

At the attribution level, Table \ref{tab:noise_interpretability} demonstrates that this structural resilience is reflected in markedly improved separation between informative and nuisance variables. CFNN-Hybrid achieves the lowest noise-to-signal attribution ratio and perfect Top-5 recovery of the ground-truth signal features, while the foundational CFNN attains the lowest mean importance rank for the true covariates. In contrast, the MLP baseline assigns nearly half as much aggregate attribution mass to nuisance features as to genuine signal features, and recovers only half of the true drivers within its top-five ranked variables. These results indicate that the CFNN family does not merely preserve predictive accuracy under feature corruption, but also maintains physically meaningful attribution structure.

\subsection{Cross-domain classification benchmarking}

To evaluate the empirical utility and generalization capability of the CFNN architectural family, we extended our benchmarking to six diverse real-world datasets spanning tabular, image, and textual modalities. The results demonstrate that the CFNN family achieves highly competitive classification accuracy relative to both standard MLPs and established continued fraction architectures. In the textual domain, CFNN-Hybrid exhibits a remarkable performance leap, achieving accuracy improvements of 4 percentage points and 6 percentage points over the state-of-the-art CoFrNet-DL on the Sentiment and Quora datasets, respectively. This significant gain is attributed to the structural innovations of parallel rational units and skip connections, which enhance sensitivity to sparse semantic features. While CFNN models encounter challenges on image-based datasets due to the absence of dedicated spatial interaction modules, they maintain a clear advantage in learning efficacy over general-purpose architectures such as MLP and CoFrNet-D. These findings underscore the robust representational capacity inherent in the CFNN architecture across heterogeneous data structures.

\subsection{Scientific interpretability and ``grey-box'' modeling}

The transition from black-box predictive modeling to transparent scientific discovery necessitates architectures that yield physically consistent feature attributions. We evaluated the CFNN family's capacity for domain-aligned representation using the UCI Energy Efficiency dataset in a single-target thermal-load forecasting setting.

\begin{figure}
  \centering
  \makebox[\textwidth][c]{%
  \includegraphics[width=1.5\linewidth]{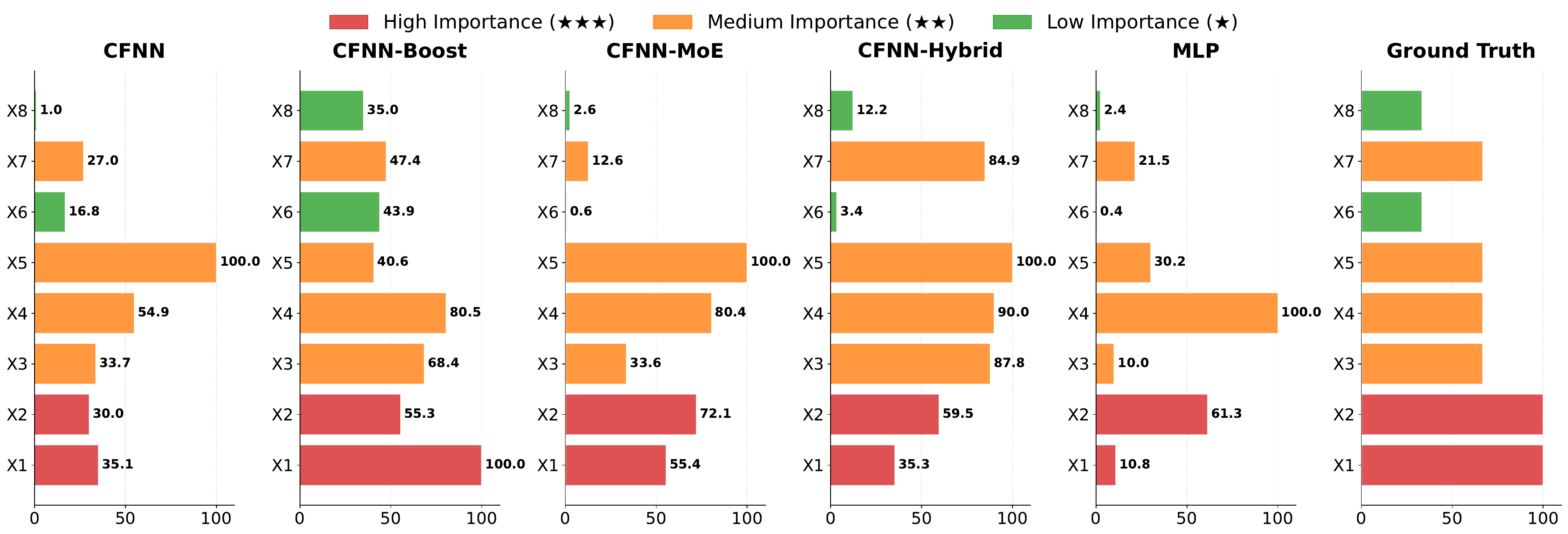}
  }
  \caption{Feature importance analysis results on the UCI Energy Efficiency dataset. Features are annotated as high, medium, or low importance based on their true relevance, with distinct colors used to indicate each importance level in the figure.}
  \label{fig:feature_importance}
\end{figure}

\begin{figure}
  \centering
  \makebox[\textwidth][c]{%
  \includegraphics[width=1.5\linewidth]{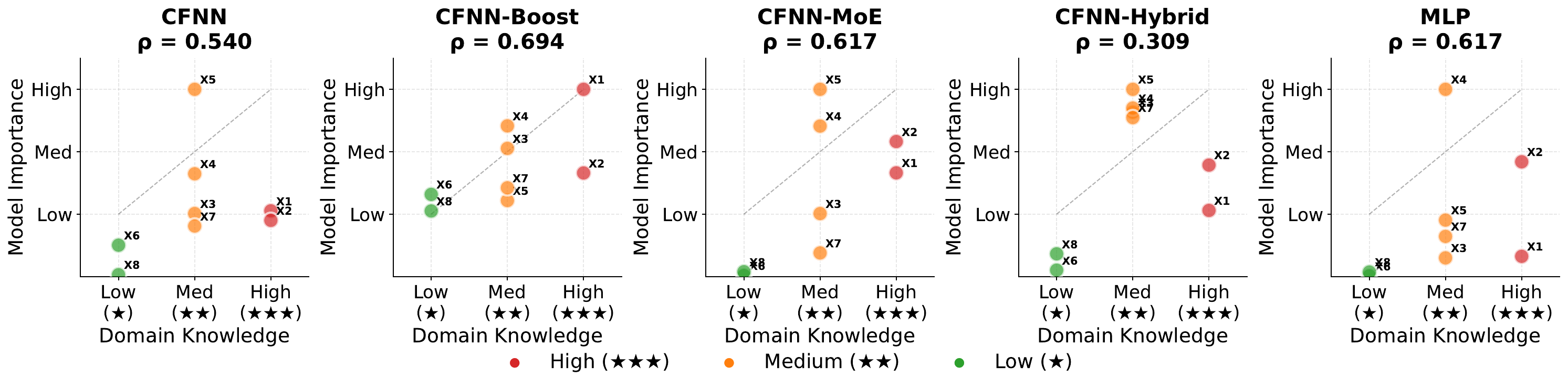}
  }
  \caption{Feature importance correlation analysis on the UCI Energy Efficiency dataset. Ideally, features within each importance category should cluster around the gray reference line, indicating accurate alignment between estimated and true importance rankings.}
  \label{fig:feature_importance_correlation}
\end{figure}

As depicted in Fig. \ref{fig:feature_importance}, SHAP-based attribution analysis reveals that CFNN variants, specifically CFNN-Boost and CFNN-Hybrid, autonomously prioritize high-relevance determinants such as 'Relative Compactness'. In contrast, the MLP baseline is susceptible to spurious correlations from variables with negligible physical impact. The fidelity of these representations is further substantiated by the correlation analysis in Fig. \ref{fig:feature_importance_correlation}, where CFNN architectures exhibit tight clustering along the ideal alignment diagonal, signifying a decision-making process rooted in thermodynamic principles. Conversely, the MLP's stochastic scatter reflects a fundamental lack of structural alignment with underlying physics. Quantitatively, Table \ref{tab:energy_performance} shows that CFNN-Hybrid achieves superior generalization with an $R^{2}=0.9317$, while Table \ref{tab:energy_interpretability} demonstrates that CFNN-Boost attains the strongest rank correlation with domain knowledge of $\rho=0.6944$ and the highest Top-3 recovery rate of 66.7\%. By comparison, the MLP baseline achieves the highest overall ranking consistency (75.0\%) but recovers only one-third of the truly important variables within its top-three features. These results establish the CFNN family as a robust ``grey-box'' paradigm that reconciles high-precision approximation with rigorous scientific transparency.

\section{Discussion}\label{sec: Discussion}

The development of the CFNN architecture family introduces a fundamental paradigm shift in function approximation by synthesizing the rigorous representation capabilities of classical continued fractions with the flexibility of modern gradient-based optimization. Our systematic evaluations demonstrate that this integration effectively addresses the ``spectral bias'' inherent in traditional MLPs, which frequently prioritize low-frequency components at the expense of local precision. While standard architectures struggle to resolve sharp transitions, singularities, or intrinsic rational structures without excessive parameter scaling, CFNN leverages its rational inductive bias to model these complex manifolds with high precision and minimal structural redundancy.

The observed Pareto dominance of the CFNN variants—most notably the CFNN-Hybrid architecture—underscores a critical structural insight: the mere expansion of hidden layer width or network depth in traditional models is insufficient to overcome the intrinsic limitations imposed by fixed activation functions. In contrast, the adaptive rational units within the CFNN framework enable the recovery of underlying functional hierarchies that remain computationally inaccessible to even heavily over-parameterized MLPs. This efficiency is particularly evident in our comparative benchmarking against state-of-the-art architectures like CoFrNet; the CFNN family not only achieves superior accuracy across diverse tasks but also establishes a new performance baseline for rational neural models by overcoming the recursive instability that has historically hindered their adoption.

Beyond raw predictive performance, CFNN successfully bridges the gap between ``black-box'' deep learning and ``white-box'' physical modeling. As evidenced by our energy efficiency case study, the architecture's mathematical abstractions do not merely fit the data but align closely with thermodynamic priorities and domain-specific hierarchies. This ``grey-box'' transparency, further reinforced by a 47-fold improvement in noise suppression efficiency, provides a robust modeling tool for scientific research where interpretability and reliability are paramount. The ability of CFNN to isolate physically significant signals from high-dimensional stochastic noise suggests that its rational structure functions as a powerful form of implicit regularization, ensuring that the learned mappings are both expressive and scientifically valid.

A critical insight emerging from our work is the complementary specialization of the three CFNN variants in addressing distinct topological challenges. CFNN-Hybrid, with its regularized denominator design, deliberately sacrifices the ability to represent true discontinuities in favor of absolute numerical stability, making it the optimal choice for fitting continuous manifolds with extreme curvature. In contrast, CFNN-Boost leverages stage-wise additive modeling to directly address genuine discontinuities and abrupt transitions, as subsequent weak learners can explicitly target the large residuals left at breakpoints by previous stages. CFNN-MoE occupies an intermediate position, employing RBF-gated expert isolation to handle local singularities within heterogeneous manifolds while maintaining greater approximation flexibility than CFNN-Hybrid through its softer regularization constraints. This architectural diversity enables practitioners to select the variant best aligned with the topological characteristics of their target function, whether continuous high-curvature features, genuine discontinuities, or localized singular behaviors.

Despite these advancements, several avenues for future exploration remain. While CFNN variants demonstrate exceptional performance on functional and tabular data, their integration into large-scale pre-trained models, such as those used in natural language processing or computer vision, requires further investigation into synergistic evolution with advanced mechanisms like Attention. Additionally, the regularization hyperparameter $\gamma$ in CFNN-Hybrid, currently set as a fixed constant to ensure numerical stability, presents an intriguing opportunity for adaptive learning. Future work may explore treating $\gamma$ as a learnable parameter or implementing layer-wise adaptive regularization strategies, potentially enabling the model to dynamically balance between approximation flexibility and numerical robustness based on local functional characteristics.

In conclusion, by establishing a rigorous foundation for high-precision fitting, the CFNN family paves the way for a more transparent, efficient, and physically consistent era of AI-driven scientific discovery.

\section{Methods}\label{sec: Methods}
The CFNN model family is founded upon the mathematical theory of continued fractions integrated with advanced machine learning techniques, thereby overcoming the inherent challenges of gradient instability and parameter scalability associated with traditional continued fraction formulations. To comprehensively characterize the capabilities of the CFNN series, we have designed a suite of experiments that provide a holistic evaluation of their performance. In this section, we systematically detail the architectural designs of the CFNN variants and outline the specific experimental protocols employed.

\subsection{The CFNN Model Architectural Family}
The core challenge of function approximation lies in designing a mathematical mapping that possesses both robust representation power and numerical stability within modern optimization frameworks \cite{Halo}. Traditional deep learning models primarily rely on power series approximations rooted in polynomial expansion logic \cite{UniversalApproximate,MLP!}. This structure often exhibits significant limitations when handling nonlinear functions characterized by local singularities or complex asymptotic behaviors \cite{NestedLearning,FreqBiasNNICML2020}. To transcend these bottlenecks at a fundamental structural level, we introduce the CFNN architectural family.

\subsubsection{The Foundational CFNN Architecture}
The theoretical foundation of CFNN stems from the theory of continued fractions in numerical analysis and its profound connection to Padé Approximant Theory \cite{PadeApproximateJMAA1961,PadeApproximateBook2006Cuyt}. In mathematical analysis, continued fractions are renowned for their superior representation efficiency, capturing local singularities and global asymptotic properties that power series expansions often struggle to resolve \cite{CFBook1992Rockett}. Specifically, an $L$-th order continued fraction is essentially a rational function whose convergence rate typically surpasses that of polynomial approximations of the same order, thereby introducing a highly efficient inductive bias into the neural network \cite{HankelBook2003Peller,ApproximateTheoryBook2019}.

\paragraph{Formal Definition and Recursive Structure}
Let $x \in \mathbb{R}^d$ denote the input vector. The output of a CFNN is generated through the hierarchical composition of a sequence of fractional terms. Formally, the foundational mapping function $f(x)$ of a CFNN follows the nested recursive structure:

\begin{equation}
f(x)=a_{0}(x)+\frac{b_{0}(x)}{a_{1}(x)+\frac{b_{1}(x)}{a_{2}(x)+\frac{b_{2}(x)}{\dots}}}
\end{equation}

In this structure, $a_{i}(x)$ and $b_{i}(x)$ represent transformation functions of the input. To further enhance the model's expressivity and its transparency as a ``grey-box'' model, we employ projection-based scalar polynomial transformations as the core components of CFNN. Directly applying multivariate polynomial expansions to high-dimensional inputs would lead to a combinatorial explosion of parameters $O(d^p)$, severely violating the parameter frugality essential for robust scientific computing.

To circumvent the curse of dimensionality, we explicitly decouple the dimensional mapping from the nonlinear polynomial expansion. For an input vector $x \in \mathbb{R}^{d}$, each fractional term first projects $x$ onto a learnable 1D scalar subspace:

\begin{equation}
z_{a,i}=w_{a,i}^{T}x+c_{a,i}, \quad z_{b,i}=w_{b,i}^{T}x+c_{b,i}
\end{equation}

where $w_{a,i}, w_{b,i} \in \mathbb{R}^d$ are learnable projection vectors, and $c_{a,i}, c_{b,i} \in \mathbb{R}$ are bias terms. Subsequently, the univariate polynomial parameterization is applied exclusively to these projected scalars:

\begin{equation}
a_{i}(x)=\sum_{k=0}^{p}\alpha_{i,k}(z_{a,i})^{k}, \quad b_{i}(x)=\sum_{k=0}^{p}\beta_{i,k}(z_{b,i})^{k}
\end{equation}

where $p$ represents the degree of the polynomial, and $\alpha_{i,k}$ and $\beta_{i,k}$ are scalar learnable parameters optimized during training. Under this design, the key hyperparameters of a CFNN are jointly constituted by the Continued Fraction Depth $L$ and the Polynomial Degree $p$.

Under this projection-based design, the parameter complexity for each fractional layer is strictly bounded to $O(d+p)$, rather than scaling exponentially. This architectural choice forces the network to discover the most physically relevant linear combinations of features before applying high-order rational approximations, fundamentally underpinning the ``parameter frugality'' observed in our Pareto frontier analysis. This polynomially parameterized CFNN is more than a black-box mapping; it establishes a transparent mapping between parameters and functional behavior. By adjusting $L$ and $p$, researchers can flexibly balance representation precision and parameter scale based on task complexity.

\paragraph{Theoretical Capacity and Approximation Bounds}
To rigorously establish the mathematical foundation underlying the superior performance of CFNNs, we provide formal approximation guarantees that quantify the relationship between model capacity and approximation error. The following theorem characterizes the exponential convergence rate of CFNN approximations as a function of the continued fraction depth $L$ and polynomial degree $p$.

\begin{theorem}[Approximation Capacity of CFNN]
Let $f: \mathbb{R}^d \to \mathbb{R}$ be a target function belonging to a class of functions with localized singularities, such as piecewise analytic functions or functions with branch points. For a CFNN with depth $L$ and polynomial degree $p$ applied to a projected 1D latent space $z = w^T x + c$, the rational approximation error $\epsilon$ satisfies:
\begin{equation}
\sup_{z \in \Omega} |f(z) - \text{CFNN}(z; L, p)| \leq \mathcal{O}(e^{-c\sqrt{Lp}})
\end{equation}
where $c > 0$ is a constant depending on the analytic region of $f$.
\end{theorem}

This theorem reveals a fundamental advantage of the rational inductive bias over polynomial-based architectures. The approximation error decays exponentially with respect to the product of depth $L$ and polynomial degree $p$, in stark contrast to standard ReLU-based MLPs, which typically achieve only polynomial decay rates of the form $\mathcal{O}((W \times D)^{-\alpha})$ for functions with non-smooth features. This exponential convergence property provides a rigorous theoretical explanation for the parameter frugality observed empirically in our Pareto frontier analysis, where CFNNs consistently achieve superior accuracy with one to two orders of magnitude fewer parameters than MLPs.

Furthermore, the projection-based parameterization strategy offers a principled mechanism for mitigating the curse of dimensionality. Since the CFNN architecture decouples the dimensional mapping from the nonlinear polynomial expansion through the projection $z = w^T x + c$, the parameter count scales as $\mathcal{O}(d+p)$ rather than $\mathcal{O}(d^p)$. When the target functional manifold possesses an effective low-dimensional subspace structure, the approximation rate becomes independent of the ambient input dimension $d$, thereby circumventing the combinatorial explosion that plagues traditional high-dimensional approximation schemes. This dimensional independence is particularly valuable in scientific computing applications where physical systems often exhibit intrinsic low-dimensional dynamics embedded in high-dimensional observation spaces.

However, while the foundational CFNN demonstrates remarkable fitting potential in small-scale tasks, its deeply nested recursive nature introduces risks of vanishing gradients and numerical instability. Consequently, we propose three evolved architectures, namely CFNN-Boost, CFNN-MoE, and CFNN-Hybrid, to address optimization bottlenecks in various scenarios.

\subsubsection{CFNN-Boost: Stage-wise Residual Refinement}

While the foundational CFNN possesses a natural advantage in representing rational structures, its deeply nested fractional construction faces severe challenges in practical optimization. As the recursive depth increases, the multi-level nesting causes the gradient flow to become highly susceptible to instability during backpropagation. This limitation restricts the model's ability to enhance capacity simply by increasing depth. To overcome this constraint and bolster parameter scalability, we introduce the CFNN-Boost architecture.

\paragraph{Motivation and Core Logic}
The core philosophy of CFNN-Boost is to transform the complexity inherent in deep nesting into an additive ensemble strategy. Rather than relying on the end-to-end optimization of a single deep network, this architecture adopts a Stage-wise Additive Modeling approach \cite{GBM!}. It constructs a powerful global mapping capability by combining multiple ``weak learners.'' Each weak learner consists of a shallow CFNN unit, thereby preserving the nonlinear fitting advantages of continued fractions while circumventing the numerical oscillation issues associated with deep recursion \cite{BoostAlgorithm2007,GBNN2020,XGBoost!}.

\paragraph{Formalization and Sequential Optimization}
Under the CFNN-Boost framework, the final predictive output of the model is expressed as a linear combination of multiple sub-models. Formally, let $F_{t-1}(x)$ represent the prediction on input $x$ by an ensemble consisting of the previous $t-1$ shallow CFNN units. At the $t$-th stage, we initialize a new shallow unit $f_t(x)$, setting its training objective to fit the residual between the ground truth label $y$ and the current ensemble prediction.

The specific optimization objective function is defined as:
\begin{equation}
\min_{\theta_t} \mathcal{L}(y, F_{t-1}(x) + \eta \cdot f_t(x; \theta_t))
\end{equation}
where $\eta \in (0, 1]$ denotes the shrinkage rate. This parameter controls the contribution of each newly added unit to the overall prediction, serving as an implicit regularizer that effectively prevents overfitting \cite{ShrinkRate2009}.

\paragraph{Architectural Characteristics and Hyperparameters}
In CFNN-Boost, the critical hyperparameters shift from a single recursive depth to the maximum number of base units and the polynomial degree within each unit. By fixing the depth of the base units at a relatively small value, CFNN-Boost can robustly enhance the overall representation ceiling of the model by increasing the number of units while maintaining controllable computational overhead.

This iterative ensemble strategy not only strengthens numerical stability but also endows CFNN-Boost with the native capability to handle functional discontinuities and abrupt transitions. By fitting residuals stage-wise, subsequent weak learners can directly target the large prediction errors left at breakpoints by previous stages, enabling effective approximation of genuinely discontinuous functions. However, additive strategies may still encounter efficiency bottlenecks in global searching when processing highly localized nonlinear manifolds. This motivates our development of the expert-partitioning-based MoE architecture.

\subsubsection{CFNN-MoE: Localized Manifold Partitioning}

While additive ensemble strategies enhance model capacity, they often struggle to reconcile all local features when mapping complex functions characterized by high heterogeneity or local singularities. To achieve refined modeling of such heterogeneous functional manifolds, we introduce the CFNN-MoE architecture. By incorporating a Mixture-of-Experts (MoE) mechanism, this architecture partitions high-dimensional feature spaces into multiple sub-manifolds \cite{MoE1991}, enabling each expert to specialize in approximating distinct local regions. This design is particularly effective for isolating and handling local singularities, as the RBF gating network can confine singular behaviors to specific expert units while maintaining smooth approximations elsewhere \cite{LocalizedReceptiveField1988}.

\paragraph{Expert Unit Design: Polynomial Priming and Backward Recursion}
Each expert unit within CFNN-MoE is not merely a simple hierarchical stack; it is constructed through a combination of polynomial priming and recursive rational structures. For a high-dimensional input $x \in \mathbb{R}^n$, the expert unit first projects it into a latent space $z = \sigma(Wx+b)$ via linear projection and nonlinear mapping. Subsequently, local feature correlations are captured using a polynomial expansion $\mathcal{P}(x;\Theta)=\sum_{j=0}^{d}c_{j}\hat{\odot}z^{j}$, enabling the expert unit to construct complex local features while maintaining an extremely low parameter count. This latent space projection design represents a natural extension of the scalar projection philosophy employed in the foundational CFNN, adapted here for complex manifold partitioning tasks where multiple expert-specific subspaces are required.

The depth $D$ of the expert unit is implemented via a unique backward recursive structure:
\begin{equation}
\begin{cases}
O_{D}(x) = \text{softplus}(\mathcal{P}_{D}(x)) + 1 \\
O_{i}(x) = \mathcal{P}_{i}(x) + \frac{\text{softplus}(\beta_{i})}{O_{i+1}(x) + \epsilon}, \quad i = D-1, \dots, 1
\end{cases}
\end{equation}
where $\beta_{i}$ represents learnable coupling strength parameters, and $\epsilon$ is a small positive constant for numerical stability. This recursive mechanism essentially approximates the Padé form of rational functions. Notably, the softer regularization constraint $O_{i+1}(x) + \epsilon$ in the denominator provides CFNN-MoE with greater flexibility in approximating local asymptotic features compared to CFNN-Hybrid's stricter squared-term regularization. This design enables the model to achieve superior convergence efficiency when handling nonlinear boundaries with sharp fluctuations or near-singular characteristics within isolated expert regions.

\paragraph{Gating Network: Anisotropic RBF-based Feature Space Partitioning}
To coordinate the specialization of various expert units, CFNN-MoE introduces a gating network based on anisotropic RBF \cite{RBF2002}, designed to achieve nonlinear partitioning of the high-dimensional feature space \cite{SparseGatedMoEICLR2017}. Standard isotropic RBF gating mechanisms assume equal importance across all input dimensions, which can lead to suboptimal performance in high-dimensional or heterogeneous feature spaces where different features exhibit vastly different scales and relevance. To address this limitation, we assign each expert $E_k$ not only a center vector $\mu_k$ and a width parameter $\sigma_k$, but also a learnable diagonal scaling matrix $\Lambda_k = \text{diag}(\lambda_{k,1}, \ldots, \lambda_{k,d})$ that adaptively weights the contribution of each feature dimension. The activation level of an input sample $x$ for a specific expert is determined by a weighted distance metric:
\begin{equation}
g_{k}(x) = \exp\left(-\frac{(x-\mu_{k})^{T} \Lambda_{k} (x-\mu_{k})}{2\sigma_{k}^{2}+\epsilon}\right)
\end{equation}
where $\lambda_{k,j} > 0$ are optimizable parameters that enable the gating network to autonomously learn feature embeddings. This anisotropic design allows the model to ignore irrelevant dimensions by driving $\lambda_{k,j} \to 0$ while stretching the manifold along critical axes, thereby effectively circumventing the curse of dimensionality in heterogeneous feature spaces. Importantly, this enhancement introduces only $d$ additional parameters per expert, maintaining alignment with the parameter frugality principle central to the CFNN philosophy. Through this localized activation mechanism, each expert is highly active only within its specific receptive field, facilitating a smooth transition from global modeling to local specialization. The final predictive output $\mathcal{Y}(x)$ is the probabilistic weighted sum of all expert outputs: $\mathcal{Y}(x) = \sum_{k=1}^{K} \pi_{k}(x) \cdot E_{k}(x)$.

\paragraph{Iterative Growth and Rollback Optimization Strategy}
Unlike traditional MoE models with fixed architectures, CFNN-MoE employs a data-driven greedy growth strategy to automatically match computational resources with task complexity. During training, the system continuously monitors the loss distribution of samples. By identifying the centroids of regions with the highest prediction errors, it determines the initial position $\mu_{K+1}$ for new expert units, ensuring that structural expansion precisely covers the most challenging manifold regions.

To prevent model redundancy and suppress overfitting, the algorithm integrates a Rollback Protection mechanism. If an additional expert unit fails to significantly improve accuracy on the validation set, the system performs a rollback operation, discarding the candidate node and locking the current optimized topology. This mechanism acts as an automatic regularizer, allowing the model to maintain an efficient scale while effectively suppressing the risk of overfitting associated with over-parameterization.

\subsubsection{CFNN-Hybrid: Parallel Rationalization and Numerical Stability}

While the nested recursive structure of continued fractions provides a powerful mathematical tool for approximating continuous functions, its inherently deep recursion still poses significant challenges in deep learning optimization \cite{ResNet2016He}. These challenges specifically involve gradient propagation efficiency and parallel computing constraints when scaling to extremely deep architectures. To address these bottlenecks, we propose CFNN-Hybrid. This architecture preserves the essence of rational approximation from continued fractions while transforming the deep recursive structure into parallel rational units \cite{RationalNNNIPS2020,NNRationalFuncICML2017}, thereby achieving a more efficient computational pipeline and a more stable gradient flow.

\paragraph{Architecture Definition and Parallelization Strategy}
Drawing inspiration from the design principles of Residual Networks \cite{ResNet2016Targ}, the core of CFNN-Hybrid is constructed by stacking a linear skip connection with $n$ parallel rational units. For a given input $x \in \mathbb{R}^d$, the final output $y$ of the network is formulated as:

\begin{equation}
y = W_{skip}x + b_{skip} + \sum_{i=1}^{n} \text{RationalUnit}_{i}(x)
\end{equation}

In this hybrid design, the linear component is responsible for capturing the global linear trends of the data, while the parallel rational units focus on learning complex nonlinear residual components. Compared to the recursive nesting of the foundational CFNN, this parallel structure significantly enhances the model's parallel computing efficiency and lays the foundation for large-scale parameter optimization.

\paragraph{Numerical Stability and Regularization Mechanism}
The most prominent issue with traditional rational functions during optimization is the risk of the denominator approaching zero, which leads to numerical singularities and training interruptions \cite{AAARationalApproximate2025}. To fundamentally resolve this, CFNN-Hybrid introduces an innovative regularization constraint mechanism for each rational unit $\mathcal{R}(x)$—defined as the ratio of polynomials $P(x)$ and $Q(x)$:

\begin{equation}
\mathcal{R}(x) = \frac{P(x)}{Q(x)^{2} + \gamma}
\end{equation}

Here, $\gamma$ is a non-negative regularization hyperparameter that governs the trade-off between approximation flexibility and numerical stability. When $\gamma = 1.0$ (the default setting in this work), the denominator is mathematically guaranteed to remain greater than or equal to 1.0 across the entire real domain, thereby eliminating poles on the real axis. This design sacrifices the ability to represent true vertical asymptotes in exchange for absolute smoothness and exceptional numerical stability, making CFNN-Hybrid particularly well-suited for fitting continuous manifolds with extreme curvature. As $\gamma \to 0$, the model gains the capacity to approximate genuine real-domain singularities, though this may introduce numerical instability in very deep networks. Therefore, in high-order scientific computing where computational efficiency and stability are paramount, we set $\gamma$ to a constant value well away from zero. Future work may explore treating $\gamma$ as a learnable parameter to enable adaptive regularization.

\paragraph{Formal Stability Analysis}
To rigorously justify the numerical stability advantages of CFNN-Hybrid, we provide a formal analysis of gradient boundedness and Lipschitz continuity. In the foundational CFNN architecture, the gradient with respect to parameters involves a chain of derivatives through the nested recursive structure. Specifically, the gradient $\nabla_\theta f(x)$ contains multiplicative terms of the form $\prod_{j} (D_j)^{-2}$, where $D_j$ denotes the denominator at the $j$-th fractional layer. When poles fall on the real axis, meaning $D_j \to 0$ for some layer $j$, this product diverges to infinity, resulting in severe gradient explosion and loss of Lipschitz continuity. This mathematical instability manifests empirically as the pronounced loss oscillations observed in Fig \ref{fig:train_loss_stability} for the foundational CFNN.

In contrast, the regularized rational parameterization of CFNN-Hybrid provides rigorous guarantees on gradient boundedness. The following lemma formalizes this stability property.

\begin{lemma}[Gradient Boundedness of CFNN-Hybrid]
For the regularized rational unit $\mathcal{R}(x) = \frac{P(x)}{Q(x)^{2} + \gamma}$ with $\gamma > 0$, the forward activation is bounded by $\sup |\mathcal{R}(x)| \leq \frac{|P(x)|}{\gamma}$, and its derivative with respect to the input $x$ satisfies:
\begin{equation}
\left| \frac{\partial \mathcal{R}(x)}{\partial x} \right| \leq C(\gamma, p)
\end{equation}
where $C(\gamma, p)$ is a constant depending on the regularization parameter $\gamma$ and polynomial degree $p$. This guarantees that the CFNN-Hybrid mapping is globally Lipschitz continuous on the real domain, explicitly eliminating the singularities of the objective function landscape.
\end{lemma}

This lemma establishes that the squared denominator regularization $Q(x)^2 + \gamma$ not only prevents numerical overflow but also ensures that all derivatives remain bounded throughout the optimization process. The global Lipschitz continuity property directly translates to the smooth training dynamics observed in our experiments, where CFNN-Hybrid exhibits markedly reduced gradient variance compared to the foundational CFNN as shown in Fig \ref{fig:train_gradient_std}. This formal stability guarantee, combined with the parallel architecture design, explains why CFNN-Hybrid achieves robust parameter scalability with monotonically decreasing approximation errors as model capacity expands.

\paragraph{Global Optimization Advantages}
As a globally optimized architecture, CFNN-Hybrid allows all learnable parameters to undergo synergistic optimization. Experimental observations indicate that, unlike the potential fluctuations in training loss seen in CFNN-Boost or CFNN-MoE when adding expert units, the training loss and test error of CFNN-Hybrid exhibit a smoother and more continuous decline. This characteristic of global optimization endows the model with more stable performance across various evaluation metrics, making it a highly competitive representative of the CFNN family in the field of scientific computing.

\subsection{Experimental Design and Evaluation Framework}\label{subsec: ExpDesign}

To comprehensively evaluate the CFNN architectural family across multiple dimensions of performance, we designed a systematic experimental framework comprising four complementary components. Each component addresses distinct aspects of model capability, from foundational approximation efficacy to practical scientific utility.

\subsubsection{Gradient Stability Analysis across Optimization Landscapes}

To quantitatively evaluate the optimization robustness of the CFNN architectural family, we designed a systematic gradient monitoring protocol. This experiment aims to identify the structural origins of the training instabilities observed in foundational continued fraction networks.

The stability benchmarks utilize two complex mathematical mappings: a composite periodic-exponential function $f(x) = \exp(x) \cdot \sin(3\pi x)$ and a high-frequency oscillatory function $f(x) = \sin(10\pi x) \cdot \exp(-x^2/10)$. These functions provide a rigorous test for gradient regularity due to their sharp transitions and dense spectral characteristics. The input domain is standardized to $[-2, 2]$, with 2,000 samples partitioned for training.

We implemented a real-time gradient tracking mechanism that hooks into all model parameters during backpropagation. Key metrics include the mean and maximum standard deviation of gradient norms, along with the frequency of numerical anomalies. To ensure comparability across diverse CFNN architectures, all models are trained for 500 epochs with a fixed learning rate of 0.001 and a gradient clipping threshold of 1.0. For the model architecture, we employ a third-order polynomial combined with a three-layer continued fraction structure to provide baseline yet intuitively interpretable results. Considering the reliability of the results, we conduct experiments using three distinct random seeds 42, 123 and 456, and report the standard deviation of the outcomes to illustrate the extent of performance variability.

\subsubsection{Structural Scalability and Depth-Performance Dynamics}

This experimental component investigates the relationship between architectural depth and representational efficacy, specifically addressing the non-monotonic scaling behavior observed in foundational CFNNs.

We systematically varied the model depth across seven discrete levels $L \in \{2, 4, 6, 8, 10, 12, 15\}$ while maintaining a constant third-order polynomial parameterization. The target mapping is a 3D exponential sum $f(x) = \exp(x_1 + x_2 + x_3)$, which requires the model to capture complex, non-separable interactions across three input dimensions.

For each depth configuration, models are trained using a depth-proportional epoch schedule to ensure sufficient convergence time for deeper architectures. We monitor test-set MSE and parameter-specific training kinetics across multiple random seeds: 42, 123, and 456. The analysis focuses on ``Scaling Efficiency'', defined as the rate of error reduction per additional parameter. This allows us to quantitatively characterize the ``critical depth'' beyond which standard recursive rational structures suffer from performance decay, and to verify the monotonic scaling robustness of the CFNN-Hybrid and MoE variants. To better benchmark the proposed architecture against recently emerging foundational model structures, we also include comparative evaluations with KAN in our scaling analysis.

To quantify the resolution of spectral bias, we performed a comprehensive frequency-domain analysis on the prediction residuals of a highly non-linear target manifold, $y = (x_1 x_2) / x_3$. This function was selected as the primary spectral benchmark due to its complex rational structure and sharp transitions that require both high-frequency resolution and low-frequency stability.

\paragraph{Spectral Evaluation Metrics}
To objectively quantify spectral bias without being confounded by the natural energy decay typical of mathematical manifolds, we utilized the Relative Power Spectral Density (Relative PSD). Let $\text{PSD}_{\text{target}}(f)$ and $\text{PSD}_{\text{residual}}(f)$ denote the power spectral densities of the ground truth signal and the model's prediction residual, respectively, obtained via Fast Fourier Transform (FFT). The Relative PSD is defined as:

\begin{equation}
Relative_{PSD}(f) = \frac{\text{PSD}_{\text{residual}}(f)}{\text{PSD}_{\text{target}}(f)}
\end{equation}

To evaluate the models' performance distribution, the frequency spectrum was partitioned into low-frequency ($f < 0.25 f_{\max}$) and high-frequency ($f \ge 0.25 f_{\max}$) regimes, where mean relative errors were calculated. Furthermore, we define the Cumulative Error Spectrum as the non-normalized integral of the Relative PSD from zero to frequency $f$:

\begin{equation}
Cumulative_{Error}(f) = \sum_{k=0}^{f} {Relative_{PSD}(k)}
\end{equation}

This cumulative metric provides a monotonic visual representation of a model's global error accumulation across the frequency domain, clearly distinguishing between frequency-biased models and frequency-agnostic architectures. The ``Spectral Advantage'' is defined as the ratio of high-frequency error power between the MLP baseline and CFNN variants with the same parameter scale, providing a quantitative measure of the model's ability to recover physically significant high-frequency signals.

\subsubsection{Lead Metric Analysis for Function Approximation Efficacy}

The foundational evaluation of CFNN's approximation capabilities employs a synthetic dataset comprising four representative functions, organized into two categories as shown in Table \ref{tab:function_list}. The first category contains the Rational Interaction and Nested Rational functions, both of which share structural similarities with continued fractions and are therefore used to assess representation consistency with the CFNN inductive bias. The second category contains the Bilinear Ratio and Runge functions, which introduce sharp transitions and nonlinear ratio structures that are well known to challenge conventional neural networks, thereby providing a stringent benchmark for approximation precision. Unless otherwise specified, all MLP baselines employed a two-hidden-layer architecture with ReLU activations. Model complexity was systematically calibrated by adjusting the number of neurons in each hidden layer, thereby enabling direct performance comparisons with CFNN variants at equivalent parameter scales.


In addition to RMSE, this study introduces a series of in-depth evaluation metrics to refine the performance analysis. To intuitively quantify the relative gains of CFNN variants over MLP baselines, we define a \textit{Lead Metric}:

\begin{equation} \label{eq:lead_metric}
\text{Lead}(\%) = \frac{\textit{Metric}_{\text{MLP}} - \textit{Metric}_{\text{CFNN}}}{\textit{Metric}_{\text{MLP}}} \times 100\%
\end{equation}

where \textit{Metric} represents training loss or test error; a positive value indicates that CFNN outperforms the MLP. For each function category, we generated 10,000 sample points using NumPy, with inputs uniformly sampled from their respective domains. The experimental protocol systematically varies model complexity by adjusting the continued fraction depth $L$ and polynomial degree $p$ across 25 distinct parameter combinations. Performance is quantified using the Lead Metric determined by Equation \ref{eq:lead_metric}, which measures relative improvement over MLP baselines at equivalent parameter scales. This analysis enables precise characterization of CFNN's sensitivity to architectural hyperparameters and identification of optimal configurations for different functional manifolds.

\subsubsection{Pareto Frontier and Training Dynamics Analysis}

To rigorously evaluate the fundamental limits of CFNN's parameter efficiency and training stability, we designed a comprehensive experimental framework comprising two complementary components: Pareto frontier analysis for parameter efficiency and training dynamics analysis for optimization stability.

The Pareto frontier is defined as the set of optimal trade-off points between model complexity and approximation accuracy. A model configuration is considered Pareto-optimal if no other configuration exists that simultaneously achieves both lower parameter count and lower approximation error. To construct these frontiers, we systematically scale model capacity across logarithmic scales by adjusting the number of parallel rational units, expert components, or ensemble members. For each architecture variant, we evaluate performance across four complex mathematical functions characterized by diverse topological structures: Jacobian elliptic functions, incomplete elliptic integrals (first and second kind), and modified Bessel functions. These functions exhibit challenging properties including double periodicity, singularities, and exponential growth characteristics that rigorously test model robustness. All functions are listed in Table \ref{tab:pareto_functions_overview}.

To examine the actual training dynamics of CFNN-Hybrid, whose superior performance was highlighted in the Pareto frontier experiments, we compare its loss convergence trajectory against that of MLPs with both equivalent and tenfold parameter counts. This comparison provides deeper insight into the capabilities of this particular variant. To evaluate the training stability and convergence characteristics of CFNN-Hybrid under maximum parameter scales, we conduct training dynamics analyses using the largest model configurations identified from the Lead Metric experiments. These configurations represent the upper bound of model complexity where architectural differences in optimization stability become most pronounced. The experimental design tracks key training metrics including training loss, validation error, gradient norms, and parameter update magnitudes across 1000 training epochs.

\subsubsection{Cross-Domain Classification Benchmarking}

To comprehensively evaluate the generalization capability and practical utility of the CFNN architectural family across diverse real-world applications, we designed a systematic cross-domain classification benchmarking framework. This experimental component assesses model performance across six standard benchmark datasets spanning multiple domains and compare with MLP and state-of-the-art continued fraction model CoFrNet family.

The dataset selection strategy ensures comprehensive coverage of diverse data modalities and task complexities. Tabular datasets: Waveform, Magic, Credit Card, test the model's ability to handle heterogeneous feature spaces and imbalanced distributions. Image classification: CIFAR10, evaluates spatial pattern recognition capabilities, For text classification tasks, we utilized the IMDb Large Movie Review Dataset (denoted as Sentiment) \cite{SentimentDataset} for binary sentiment polarity classification, and the Quora Insincere Questions Classification dataset (denoted as Quora) \cite{QuoraDataset} to assess the identification of toxic or misleading content. All datasets follow a standardized 65\%:5\%:30\% split for training, validation, and testing, with fixed random seeds to ensure reproducibility.

To adapt variable-length textual sequences for the feed-forward architecture of the CFNN family, we employed pre-trained word embeddings coupled with a pooling strategy. Specifically, raw text was tokenized, and each token was mapped to a dense semantic space using pre-trained GloVe embeddings \cite{GloVe}. To construct a fixed-dimensional feature representation for each text sample, we applied mean-pooling across the token sequence. This encoding strategy explicitly marginalizes the sequential length, distilling the sentence-level semantics into a static, fixed-dimensional dense vector that is natively compatible with our parallel rational units and stage-wise mechanisms.

For rigorous benchmarking, we compare CFNN variants against three established baselines: (1) Standard MLPs as universal approximators, (2) CoFrNet-D as a representative continued fraction architecture with deep recursive structure, and (3) CoFrNet-DL as a state-of-the-art continued fraction model with specialized local feature extraction modules. Performance metrics for CoFrNet baselines are cited directly from their original studies to ensure consistency. All models are evaluated under equivalent computational constraints using a single NVIDIA RTX 3080 GPU.

Crucially, to ensure a strictly fair evaluation of architectural inductive biases, all evaluated models, including standard MLPs, CoFrNet baselines, and all CFNN variants, were subjected to identical featurization constraints. They all received the exact same GloVe-pooled static vectors as input. No sequence-specific modules were introduced into any of the baselines. This rigorous control isolates the performance variance entirely to the structural superiority of the rational inductive bias over the standard polynomial inductive bias when processing dense semantic representations.

Model performance is quantified using classification accuracy as the primary metric, with complementary analysis of training convergence characteristics and computational efficiency. To account for stochastic variability, each experiment is repeated 5 times with different random seeds, and results are reported as mean ± standard deviation. Statistical significance is assessed using paired t-tests with Bonferroni correction for multiple comparisons. This comprehensive evaluation framework enables systematic assessment of CFNN's cross-domain generalization capabilities and identification of architectural strengths across different data modalities.

\subsubsection{Implicit Regularization and Noise Suppression Evaluation}

The evaluation protocol focuses on the structural resilience of neural architectures when isolating physically significant signals embedded within high dimensional spaces contaminated by varying intensities of feature noise.

Synthetic datasets are generated through a rigorous mathematical strategy to ensure precise attribution and reproducibility. The underlying physical signal is governed by four significant covariates sampled from a standard normal distribution $x_{true} \in \mathbb{R}^4$ with the target variable determined by the non-linear relationship $y = \sin(x_1) + \cos(x_2) + x_3 x_4 + \epsilon$ where $\epsilon$ represents Gaussian noise with a standard deviation of 0.05. To simulate diverse data corruption scenarios, three distinct categories of interference features are integrated into the input space. Redundant features are constructed through linear combinations of the significant covariates supplemented by infinitesimal perturbations. Deceptive features are engineered to exhibit a spurious Pearson correlation of 0.8 with the target variable in the training set which vanishes entirely in the testing phase to test model robustness against false associations. Purely random noise features are dynamically generated following a target noise proportion $\alpha$ to populate the remaining input dimensions according to the ratio of uninformative variables to the total feature count. The experimental corpus consists of 5000 samples partitioned into training, validation, and testing sets using a 65\% 5\% 30\% distribution with fixed stochastic seeds.

Model assessment prioritizes predictive accuracy and architectural efficiency across logarithmic parameter scales. The mean square error is utilized as the primary metric for terminal fitting precision. To characterize the fundamental limits of the performance complexity trade off, we employ Pareto frontier analysis to identify the optimal boundary for each architecture. Efficiency is further quantified by determining the minimum parameter overhead required to achieve a predefined accuracy threshold specifically a mean square error below 0.01. This framework enables a direct comparison of how different inductive biases facilitate the suppression of stochastic interference while maintaining signal integrity.

To quantify attribution robustness on this synthetic noise benchmark, we compute SHAP-based global feature importance scores for every model and derive four complementary summary statistics. The noise-to-signal ratio (NSR) is defined as the ratio between the total absolute attribution assigned to nuisance features and the total absolute attribution assigned to the four ground-truth signal covariates; lower values indicate stronger suppression of irrelevant variables. Top-5 accuracy measures the proportion of the four true signal features recovered within the five highest-ranked features. The mean importance rank (MIR) is defined as the average global attribution rank of the four true signal variables, so lower values indicate that the physically meaningful covariates are concentrated nearer the top of the ranking. Finally, Suppression is reported as the NSR normalized by the CFNN-Hybrid value, providing a direct relative measure of the attribution penalty incurred by alternative architectures.

The benchmarking suite involves a systematic evaluation of six architectural paradigms across five noise intensities ranging from 20\% to 90\% and six target parameter levels from 50 to 1200. Baselines consist of MLP utilizing tanh activation functions and KAN based on learnable spline basis functions. The CFNN family is represented by its four specialized implementations.

For the fixed-configuration predictive comparison reported in Table \ref{tab:noise_predictive}, we instantiate a representative high-redundancy setting with 5000 samples, observation noise standard deviation 0.05, and a 65\%:5\%:30\% train-validation-test split using random seed 42. The input space contains four true signal features, four random noise features, two redundant features perturbed with Gaussian noise of standard deviation 0.05, and one deceptive feature constructed to achieve a training-set correlation of 0.8 with the target. All models are optimized with learning rate 0.001, batch size 64, early-stopping patience 15, and a maximum of 50 epochs. Architectural configurations are fixed as follows: foundational CFNN with depth 4 and polynomial degree 4; CFNN-Hybrid with unit degree 4 and 20 parallel rational units; CFNN-Boost with shallow depth 4, polynomial degree 4, 5 boosting stages, and 20 epochs per stage; and CFNN-MoE with shallow depth 2 per expert, polynomial degree 2, and 5 experts.

\subsubsection{Feature Importance Analysis for Scientific Interpretability}

The transition from ``black-box'' machine learning models to ``grey-box'' scientific tools requires architectures that not only achieve high predictive accuracy but also provide physically consistent feature attributions. In scientific computing applications, model interpretability is paramount, researchers must understand which input variables drive predictions and whether these attributions align with established domain knowledge. The CFNN architecture, with its mathematically transparent continued fraction structure, is uniquely positioned to bridge this gap by offering both expressive power and interpretable feature importance.

To evaluate CFNN's capacity for scientifically meaningful feature attribution, we employ the UCI Energy Efficiency dataset for building thermal load prediction. This dataset comprises 768 samples with eight input features that characterize building geometry, orientation, and thermal properties, alongside two original output variables representing heating and cooling loads. In this study, all predictive and attribution experiments on this dataset are conducted in a single-target thermal-load regression setting, such that the performance metrics in Table \ref{tab:energy_performance} and the attribution analyses in Table \ref{tab:energy_interpretability} are reported for the same target variable. The physical significance of each feature is well-established through thermodynamic principles: relative compactness, surface area, wall area, roof area, overall height, orientation, glazing area, and glazing area distribution collectively determine thermal energy requirements. This domain knowledge provides ground-truth physical hierarchies against which model attributions can be quantitatively compared.

Model interpretability is assessed using SHAP, a game-theoretic approach that provides consistent and locally accurate feature importance values. For each CFNN variant and baseline model, we compute SHAP values across the entire dataset, quantifying the marginal contribution of each feature to individual predictions. These attributions are aggregated to produce global feature importance rankings that reflect each variable's overall influence on thermal load predictions.

To rigorously assess alignment between model attributions and physical principles, we employ three complementary metrics:

\textbf{Spearman Rank Correlation Coefficient ($\rho$):} Measures the monotonic relationship between model-derived feature importance rankings and ground-truth physical hierarchies established through domain expertise. Higher $\rho$ values indicate stronger agreement with physical principles, with $\rho = 1$ representing perfect rank correlation. \textbf{Ranking Consistency:} Quantifies the percentage agreement between model rankings and physical priorities for key determinant pairs. Specifically, we evaluate whether the model correctly identifies that relative compactness $X_1$ and surface area $X_2$ should dominate thermal load predictions over less influential variables like orientation $X_6$. \textbf{Top-$k$ Accuracy:} Assesses the model's ability to identify the $k$ most physically significant features. We evaluate Top-3 accuracy to identify the three primary determinants of thermal load and Top-5 accuracy, providing granular insight into feature selection precision.

To ensure robustness, each analysis is repeated across 10 different random train-test splits by 65\%:5\%:30\%, with results reported as mean ± standard deviation. Statistical significance of differences between CFNN variants and baselines is assessed using paired t-tests with Bonferroni correction. This comprehensive evaluation framework enables systematic assessment of CFNN's capacity to provide physically consistent interpretations while maintaining high predictive accuracy in data-constrained scientific applications.

\newpage


\bibliography{sn-bibliography.bib}

\newpage

\section{Tables}\label{sec: Tables}

\begin{table}[h]
\centering
\caption{Synthetic benchmark functions used to evaluate function approximation performance. The four functions are organized into two task categories: continued-fraction-aligned functions, used to assess whether the CFNN inductive bias matches nested rational structure, and conventional-neural-network-challenging functions, used to test approximation accuracy on manifolds with sharp transitions or nonlinear ratio interactions. Here, $\epsilon$ denotes a small positive constant introduced to avoid numerical singularity in the denominator.}
\label{tab:function_list}
\begin{tabular}{lll}
\toprule
Function & Expression & Task category \\
\midrule
Bilinear Ratio Function & $\frac{x_1 x_2}{x_3+\epsilon}$ & Conventional-NN-challenging \\
Runge Function & $\frac{1}{1+25x^2}$ & Conventional-NN-challenging \\
Rational Interaction Function & $\frac{x_1+x_2}{1+x_3^2}$ & Continued-fraction-aligned \\
Nested Rational Function & $x_1+\frac{1}{x_2+\frac{1}{x_3}}$ & Continued-fraction-aligned \\
\bottomrule
\end{tabular}
\end{table}

\begin{table}[h]
\centering
\caption{RMSE for function approximation on the four synthetic benchmark functions at a matched model scale of approximately 200 learnable parameters. Lower values indicate better performance. The best result among the MLP and CFNN-family models is highlighted in bold; KAN is reported separately as an external reference baseline.}
\label{tab:basic_function_fit_exp}
\begin{tabular}{lcccc}
\toprule
\textbf{Model} & \textbf{Bilinear Ratio} & \textbf{Runge}   & \textbf{Rational Interaction} & \textbf{Nested Rational} \\
\midrule
MLP            & 1.08                    & 4.05e-1          & 1.34                          & 8.65e-1                  \\
CFNN           & 1.22                    & 4.53e-1          & 1.19                          & 5.51e-1                  \\
CFNN-Boost     & 1.97e-1                 & \textbf{4.97e-3} & 3.96e-2                       & 2.12e-2                  \\
CFNN-MoE       & \textbf{2.04e-2}        & 5.98e-3          & \textbf{3.08e-2}              & \textbf{1.98e-2}         \\
CFNN-Hybrid    & 2.27e-2                 & 5.09e-3          & 3.09e-2                       & 2.07e-2                  \\
\midrule
KAN            & 2.18e-3                 & 2.03e-3          & 6.63e-3                       & 9.8e-5                   \\
\bottomrule
\end{tabular}
\end{table}

\begin{table}[h]
\centering
\caption{Classification accuracy across six real-world benchmark datasets spanning tabular, image, and text modalities. Higher values indicate better performance. Results are reported under the standardized train-validation-test protocol described in Methods. The best result in each dataset is highlighted in bold; tied best results are both highlighted.}
\begin{tabular}{lcccccc}
\toprule
\textbf{Model} & \textbf{Waveform} & \textbf{Magic} & \textbf{Credit Card} & \textbf{CIFAR10} & \textbf{Sentiment} & \textbf{Quora} \\
\midrule
CoFrNet-DL   & 0.87 & 0.86 & 0.71 & \textbf{0.87} & 0.84 & 0.88 \\
CoFrNet-D    & 0.69 & 0.76 & 0.66 & 0.38 & 0.80 & 0.75 \\
MLP          & 0.34 & 0.65 & 0.50 & 0.35 & 0.83 & 0.85 \\
\midrule
CFNN         & 0.84 & 0.85 & \textbf{0.82} & 0.44 & 0.80 & 0.92 \\
CFNN-Boost   & 0.88 & \textbf{0.88} & \textbf{0.82} & 0.44 & 0.82 & 0.88 \\
CFNN-MoE     & \textbf{0.89} & 0.87 & \textbf{0.82} & 0.45 & 0.87 & \textbf{0.94} \\
CFNN-Hybrid  & 0.84 & 0.87 & \textbf{0.82} & 0.47 & \textbf{0.88} & \textbf{0.94} \\
\bottomrule
\end{tabular}
\label{tab:classification_exp}
\end{table}

\begin{table}[h]
\centering
\caption{Complex analytical benchmark functions used in the Pareto frontier and training-dynamics experiments. These four two-variable function families were selected to probe complementary challenges for approximation, including double periodicity, near-singular integral structure, and rapidly varying special-function behaviour. Together, they provide a stringent test bed for evaluating the trade-off between parameter count, fitting accuracy, and optimization stability across model architectures.}
\label{tab:pareto_functions_overview}
\begin{tabular}{lll}
\toprule
\textbf{Function family} & \textbf{Notation / definition} & \textbf{Input domain $\mathcal{D}$} \\
\midrule
Jacobian elliptic function $\mathrm{sn}$ & $\mathrm{sn}(x \mid y)$ & $x \in [-5, 5], y \in [0, 1]$ \\
Incomplete elliptic integral of the first kind & $F(x,y) = \int_0^x \frac{dt}{\sqrt{1 - y \sin^2 t}}$ & $x \in [0, 2\pi], y \in [0, 1]$ \\
Incomplete elliptic integral of the second kind & $E(x,y) = \int_0^x \sqrt{1 - y \sin^2 t} \, dt$ & $x \in [0, 2\pi], y \in [0, 1]$ \\
Modified Bessel function of the first kind & $I_x(y)$ & $x \in [0, 5], y \in [0, 5]$ \\
\bottomrule
\end{tabular}
\end{table}

\begin{table}[h]
\centering
\caption{SHAP-based evaluation of noise suppression and attribution fidelity on the synthetic feature-noise benchmark. NSR denotes the noise-to-signal attribution ratio, computed as the total absolute attribution assigned to nuisance features divided by that assigned to the four ground-truth signal covariates. Top-5 accuracy reports the fraction of true signal features recovered within the five highest-ranked features. MIR denotes the mean importance rank of the four true signal variables. Suppression reports NSR normalized by the CFNN-Hybrid value. Lower NSR, MIR, and Suppression are better; higher Top-5 accuracy is better. The best result in each metric is highlighted in bold; tied best results are all highlighted.}
\label{tab:noise_interpretability}
\begin{tabular}{lcccc}
\toprule
Model & NSR ($\downarrow$) & Top-5 accuracy ($\uparrow$) & MIR ($\downarrow$) & Suppression ($\downarrow$) \\
\midrule
MLP & 0.474 & 50.0\% & 6.2 & 47.4$\times$ \\
\textbf{CFNN} & 0.019 & \textbf{100.0\%} & \textbf{2.5} & 1.9$\times$ \\
CFNN-Boost & 0.071 & \textbf{100.0\%} & 2.8 & 7.1$\times$ \\
CFNN-MoE & 0.067 & \textbf{100.0\%} & 3.2 & 6.7$\times$ \\
CFNN-Hybrid & \textbf{0.010} & \textbf{100.0\%} & 3.0 & \textbf{1.0$\times$} \\
\bottomrule
\end{tabular}
\end{table}

\begin{table}[h]
\centering
\caption{Minimum parameter count required for each model to achieve prediction error below the threshold $MSE < 0.01$ under increasing levels of noise-feature interference. Parameter values are reported over the discrete search grid described in Methods. Lower values indicate greater parameter efficiency. The best result at each noise ratio is highlighted in bold; tied best results are both highlighted. A dash indicates that the model did not reach the threshold within the evaluated parameter range.}
\label{tab:noise_threshold_params}
\begin{tabular}{lcccccc}
\toprule
\textbf{Noise ratio} & \textbf{MLP} & \textbf{KAN} & \textbf{CFNN} & \textbf{CFNN-Boost} & \textbf{CFNN-MoE} & \textbf{CFNN-Hybrid} \\
\midrule
20\% & 800 & 400 & \textbf{50} & 400 & 400 & \textbf{50} \\
40\% & 800 & 1200 & \textbf{50} & 400 & 800 & 100 \\
60\% & 1200 & - & 100 & - & 1200 & \textbf{50} \\
80\% & 800 & 1200 & 200 & 1200 & 400 & \textbf{50} \\
90\% & 800 & - & \textbf{100} & - & 400 & 400 \\
\bottomrule
\end{tabular}
\end{table}

\begin{table}[h]
\centering
\caption{Predictive performance under the fixed high-dimensional feature-redundancy configuration described in Methods. Lower MSE, MAE, and Relative MSE indicate better predictive accuracy; higher $R^2$ indicates better goodness of fit. Relative MSE is reported with respect to the CFNN-Hybrid value. The best result in each metric is highlighted in bold.}
\label{tab:noise_predictive}
\begin{tabular}{lcccc}
\toprule
Model & MSE ($\downarrow$) & MAE ($\downarrow$) & $R^2$ ($\uparrow$) & Relative MSE ($\downarrow$) \\
\midrule
MLP & 0.3037 & 0.4179 & 0.6668 & 63.3$\times$ \\
CFNN & 0.0087 & 0.0717 & 0.9904 & 1.8$\times$ \\
CFNN-Boost & 0.0296 & 0.0843 & 0.9214 & 6.2$\times$ \\
CFNN-MoE & 0.0164 & 0.0954 & 0.9820 & 3.4$\times$ \\
CFNN-Hybrid & \textbf{0.0048} & \textbf{0.0553} & \textbf{0.9948} & \textbf{1.0$\times$} \\
\bottomrule
\end{tabular}
\end{table}

\begin{table}[h]
\centering
\caption{Predictive performance on the UCI Energy Efficiency dataset under the single-target thermal-load regression setting described in Methods. Lower MSE and MAE indicate smaller prediction errors, whereas higher $R^2$ indicates better goodness of fit. The best result in each metric is highlighted in bold.}
\label{tab:energy_performance}
\begin{tabular}{lccc}
\toprule
Model & MSE ($\downarrow$) & MAE ($\downarrow$) & $R^2$ ($\uparrow$) \\
\midrule
MLP & 411.60 & 16.56 & -2.9449 \\
CFNN & 61.64 & 5.85 & 0.4092 \\
CFNN-Boost & 64.01 & 5.99 & 0.3865 \\
CFNN-MoE & 81.16 & 6.03 & 0.4829 \\
CFNN-Hybrid & \textbf{7.13} & \textbf{1.93} & \textbf{0.9317} \\
\bottomrule
\end{tabular}
\end{table}

\begin{table}[h]
\centering
\caption{Alignment between model-derived feature-importance rankings and thermodynamic domain knowledge on the UCI Energy Efficiency dataset. Spearman $\rho$ quantifies rank correlation with the ground-truth physical hierarchy, Ranking Consistency reports the overall agreement between the estimated and domain-informed orderings, and Top-3 Accuracy measures the fraction of truly important variables recovered among the three highest-ranked features. Higher values indicate better interpretability alignment. The best result in each metric is highlighted in bold.}
\label{tab:energy_interpretability}
\begin{tabular}{lccc}
\toprule
Model & Spearman $\rho$ ($\uparrow$) & Ranking Consistency ($\uparrow$) & Top-3 Accuracy ($\uparrow$) \\
\midrule
Standard CFNN & 0.5401 & 64.3\% & 33.3\% \\
Hybrid CFNN & 0.3086 & 60.7\% & 33.3\% \\
Boost CFNN & \textbf{0.6944} & 71.4\% & \textbf{66.7\%} \\
MLP & 0.6172 & \textbf{75.0\%} & 33.3\% \\
\bottomrule
\end{tabular}
\end{table}

\clearpage

\section{Figures}\label{sec: Figures}

\backmatter

\end{document}